\documentclass[10pt,twocolumn,letterpaper]{article}

\usepackage[pagenumbers]{cvpr}

\usepackage[accsupp]{axessibility} % Improves PDF readability for those with visual impairments.
\usepackage{url}            %
\usepackage{booktabs}       %
\usepackage{amsfonts}       %
\usepackage{nicefrac}       %
\usepackage{graphicx}
\usepackage{xcolor,colortbl}         %
\usepackage[inline]{enumitem}
\usepackage{amsmath}   %
\usepackage{xspace}
\usepackage{makecell}  %
\usepackage{amssymb}   %
\usepackage{pifont}    %
\usepackage{bbm}
\usepackage{multirow}
\usepackage{ifthen}
\usepackage{alphalph}

\newcommand{\methodshort}{\textsc{SynViTA}\xspace}
\newcommand{\methodfull}{\textsc{Synthetic videos for Video-Text Alignment}\xspace}

\definecolor{VideoConBlue}{RGB}{218,232,252}
\definecolor{RetrievalRed}{RGB}{248,206,204}
\definecolor{VideoQAGreen}{RGB}{213,232,212}
\definecolor{forestgreen(web)}{rgb}{0.13, 0.55, 0.13}

\newcommand{\inlineColorbox}[2]{\begingroup\setlength{\fboxsep}{1pt}\colorbox{#1}{\hspace*{2pt}\vphantom{Ay}#2\hspace*{2pt}}\endgroup}

\newcommand{\PreserveBackslash}[1]{\let\temp=\\#1\let\\=\temp}
\newcolumntype{C}[1]{>{\PreserveBackslash\centering}p{#1}}
\newcolumntype{b}{>{\columncolor{VideoConBlue}}c|}
\newcolumntype{g}{>{\columncolor{VideoQAGreen}}c|}
\newcolumntype{r}{>{\columncolor{RetrievalRed}}c|}

\usepackage{xspace}
\makeatletter
\DeclareRobustCommand\onedot{\futurelet\@let@token\@onedot}
\def\@onedot{\ifx\@let@token.\else.\null\fi\xspace}

\def\eg{\emph{e.g}\onedot} 
\def\ie{\emph{i.e}\onedot}

\def\wrt{w.r.t\onedot}

\definecolor{darkgreen}{rgb}{0.0, 0.5, 0.0}
\newcommand{\ind}[1]{\footnotesize{\textcolor{darkgreen}{($\uparrow$ #1)}}}
\newcommand{\decd}[1]{\footnotesize{\textcolor{red}{($\downarrow$ #1)}}}

\makeatother

\newcommand{\suppmat}{Supp. Mat.}

\newcommand{\Real}{\mathbb{R}}

\newcommand{\video}{{V}}
\newcommand{\videospace}{\mathcal{V}}

\newcommand{\noisespace}{\mathcal{N}}
\newcommand{\videoreal}{\video^{r}}
\newcommand{\videosynth}{\video^{s}}

\newcommand{\textinput}{{t}}
\newcommand{\word}{\mathbf{w}}
\newcommand{\textspace}{\mathcal{T}}
\newcommand{\textreal}{\textinput^r}
\newcommand{\textsynth}{\textinput^s}

\newcommand{\textmasked}{\textinput'}

\newcommand{\loss}{\mathcal{L}}
\newcommand{\losssynth}{\loss_{\mathtt{syn}}}
\newcommand{\lossconsis}{\loss_{\mathtt{scr}}}
\newcommand{\lossreal}{\loss_{\mathtt{real}}}
\newcommand{\lambdaconsis}{\lambda_{{\mathtt{scr}}}}

\newcommand{\words}{\mathcal{W}}
\newcommand{\sharedspace}{\Real^d}%

\newcommand{\wordssimplex}{\Delta^{|\words|}}
\newcommand{\targetfunction}{f}
\newcommand{\visualencoder}{f_{\mathtt{vid}}}
\newcommand{\textencoder}{f_{\mathtt{txt}}}
\newcommand{\textdecoder}{f_{\mathtt{dec}}}

\newcommand{\generator}{G}
\newcommand{\weight}{\omega}
\newcommand{\noise}{\eta}

\newcommand{\probyes}{P_\words(\texttt{Yes}|\video,\textinput)}
\newcommand{\probno}{P_\words(\texttt{No}|\video,\textinput)}
\newcommand{\prob}{P_\words(\word|\video,\textinput)}

\newcommand{\videocon}{VideoCon\xspace}

\newcommand{\mplugowl}{mPLUG-Owl 7B\xspace}
\newcommand{\videollava}{Video-LLaVA\xspace}

\newcommand{\pali}{End-to-End VNLI\xspace}

\newcommand{\instructblip}{InstructBLIP\xspace}

\newcommand{\llava}{LLaVA-1.5\xspace}

\newcommand{\clipflantfive}{CLIP-FlanT5\xspace}

\newcommand{\cogvideox}{CogVideoX\xspace}

\newcommand{\lavie}{LaVie\xspace}

\newcommand{\videocrafter}{VideoCrafter2\xspace}

\definecolor{cvprblue}{rgb}{0.21,0.49,0.74}
\usepackage[pagebackref,breaklinks,colorlinks,allcolors=cvprblue]{hyperref}

\def\paperID{16494} %
\def\confName{CVPR}
\def\confYear{2025}

\title{Can Text-to-Video Generation help Video-Language Alignment?}

\author{
{Luca Zanella\textsuperscript{1} 
\quad 
Massimiliano Mancini\textsuperscript{1} 
\quad Willi Menapace\textsuperscript{2}}\\\vspace{-1cm}
\and
{Sergey Tulyakov\textsuperscript{2} 
\quad 
Yiming Wang\textsuperscript{3} 
\quad Elisa Ricci\textsuperscript{1,3}}\\
{\small \textsuperscript{1}University of Trento \quad \textsuperscript{2} Snap Inc. \quad \textsuperscript{3}Fondazione Bruno Kessler}\\[1mm]
{\tt\small \url{https://lucazanella.github.io/synvita/}}
}

\begin{document}
\maketitle
\begin{abstract}
Recent video-language alignment models are trained on sets of videos, each with an associated positive caption and a negative caption generated by large language models.
A problem with this procedure is that negative captions may introduce linguistic biases, \ie, concepts are seen only as negatives and never associated with a video. While a solution would be to collect videos for the negative captions, existing databases lack the fine-grained variations needed to cover all possible negatives. In this work, we study whether synthetic videos can help to overcome this issue. Our preliminary analysis with multiple generators shows that, while promising on some tasks, synthetic videos harm the performance of the model on others. We hypothesize this issue is linked to noise (semantic and visual) in the generated videos and develop a method, \methodshort, that accounts for those. \methodshort dynamically weights the contribution of each synthetic video based on how similar its target caption is \wrt the real counterpart. Moreover, a semantic consistency loss makes the model focus on fine-grained differences across captions, rather than differences in video appearance. 
Experiments show that, on average, \methodshort improves over existing methods on VideoCon test sets and SSv2-Temporal, SSv2-Events, and ATP-Hard benchmarks, being a first promising step for using synthetic videos when learning video-language models.

\end{abstract}

\section{Introduction} 
\label{sec:intro}

 \begin{figure}[t]
    \centering
    \includegraphics[width=1\linewidth]{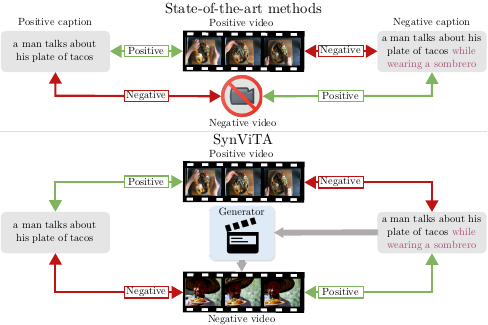}
    \caption{%
    We study the problem of video-language alignment, %
    \ie, modeling the relationship between video content and text descriptions. Top: current methods use LLM-generated negative captions, which may introduce certain concepts (\eg, \textit{wearing a sombrero}) only as negatives, as they are not associated with any video. %
    Bottom: we study whether overcoming this issue by pairing negative captions with generated videos can improve VLA.}
    \label{fig:teaser}
\end{figure}

Video-language alignment~(VLA) aims to model the relationship between video content and natural language descriptions \cite{xu2021videoclip}, a fundamental multimodal task that enables various applications, such as video captioning~\cite{fei2024tpami} and video-text retrieval~\cite{ventura2024covr}. This task is challenging because it requires the models to recognize not only the entities but also their spatial and temporal relationships. 

Recent approaches 
exploit multimodal large language models (MLLMs) to address VLA \cite{bansal2024videocon,lin2023revisiting,lin2024evaluating} by tasking the MLLM to answer whether a given video and description are aligned.
While effective, such MLLMs often lack sufficient understanding of temporal dynamics, such as action types or temporal orders
\cite{liu2024tempcompass, cores2024tvbench}.
This limitation also stems from the video-and-language datasets used for the MLLM pre-training, as they are biased towards frame-level semantics: the appearance of a single frame is often sufficient to infer the alignment with the textual caption~\cite{lei2023revealing, momeni2023verbs}. 

While a possible solution is to augment datasets with negative captions, \eg, captions from other videos, these negatives can be ``easy'' for VLA models to distinguish simply by focusing on nouns; therefore, recent works focus on LLM-generated captions as hard negatives~\cite{bansal2024videocon}. 
However, relying solely on textual negatives may cause the MLLM to encounter concepts only as negatives, %
thus developing incorrect linguistic biases. For instance, in the VideoCon dataset \cite{bansal2024videocon}, words like \textit{sombrero}, \textit{marshmallow}, and \textit{bland} appear as textual negatives but not as positives. While a remedy is to augment the training set with videos corresponding to the hard negative captions, retrieving such videos from existing databases is not a feasible solution as they lack sufficient videos that vary only \wrt actions or temporal order while remaining similar in all the other semantic aspects \cite{momeni2023verbs}. An alternative pathway is generating synthetic videos by feeding %
hard negative captions to text-to-video generative models~\cite{yang2024cogvideox,wang2023lavie,chen2024videocrafter2,menapace2024snap}. 
While this idea has been investigated in the image domain~\cite{patel2024tripletclip}, it remains largely unexplored for videos.

In this paper, we aim to fill this gap and investigate, for the first time, the use of synthetic videos to improve VLA in temporal understanding. 
Specifically, we propose to leverage negative captions generated by existing models~\cite{bansal2024videocon} and recent open-source text-to-video generators~\cite{yang2024cogvideox,wang2023lavie,chen2024videocrafter2} to produce the corresponding synthetic videos (see \cref{fig:teaser}). 
We first conduct a preliminary study to evaluate whether these generated videos can augment the training set of real videos and enhance performance on various video-related tasks. Our analysis shows that, while adding synthetic videos shows some promise, \textit{it does not} consistently improve performance on temporally challenging downstream tasks, regardless of the generator.
We also analyze the effects of different misalignment types (\ie, semantically plausible changes in the video captions) on the generated videos. We notice that videos generated by, \eg, introducing hallucination into the captions or reversing event order, align more with positive captions than with their target captions. Such noisy supervision signals may lead to ineffective learning, limiting improvements on downstream tasks.

Motivated by these preliminary findings, we argue that, when using synthetic videos for VLA we should account for (i) potential semantic inconsistency between input text and the generated videos and (ii) appearance biases, as synthetic videos may contain artifacts. We design \methodfull (\methodshort), a model-agnostic method that can effectively tackle both challenges. 
\methodshort addresses the {semantic inconsistency} problem by making the contribution of each synthetic video in the training objective proportional to their video-text alignment estimates~\cite{lin2024evaluating}.  
Moreover, it accounts for appearance biases via a semantic regularization objective that (i) takes the common parts between the original and negative caption; (ii) encourages the model to focus on semantic changes rather than on the visual appearance difference between synthetic and real videos.
We evaluate \methodshort~on the VideoCon~\cite{bansal2024videocon} test sets with different Video LLMs~\cite{ye2023mplug,lin2023video}, and on temporally challenging downstream tasks, \ie, text-to-video retrieval on SSv2-Temporal \cite{sevilla2021only} and SSv2-Events \cite{bagad2023test} and video question answering on ATP-Hard \cite{buch2022revisiting}. On average, \methodshort improves over state-of-the-art methods that do not use synthetic videos, demonstrating that synthetic videos can help VLA.

\noindent\textbf{Contributions.} 
To summarize, our contributions are:
\begin{itemize}
   \item We pioneer the research problem in how to effectively leverage synthetic videos for VLA learning to improve temporal understanding;
   \item We conduct extensive analysis, shedding light on the potential benefits and limitations of using videos generated by state-of-the-art text-to-video generative models;
   \item We propose a new learning method for VLA with synthetic videos, \methodshort, with a sample weighting strategy to mitigate noisy generations and a regularization term to enforce semantic understanding, instead of visual differences between synthetic and real videos. 
   \item We evaluate \methodshort~on different benchmarks with different Video LLMs, proving its model-agnostic effectiveness in aiding VLA for better temporal understanding.
\end{itemize}

\section{Related work} 
\label{sec:related}

\noindent\textbf{Video-language models for video understanding.} 
Recent approaches for video understanding exploit the capabilities of foundation models. For instance, several works adapted CLIP \cite{radford2021learning}, a model trained to compare images and texts, for video-language tasks, such as retrieval \cite{fang2021clip2video}, captioning \cite{luo2022clip4clip} or anomaly detection \cite{zanella2023delving}. Other studies leveraged LLMs for reasoning over video captions \cite{zanella2024harnessing,wang2022language} or directly decode video features in natural language \cite{xu2023mplug,zhang2023video,lin2023video}. 
While these models heavily rely on pre-training on large-scale video-text pairs \cite{xu2021videoclip,xu2023mplug}, they still lack robustness in modeling temporal dynamics %
\cite{liu2024tempcompass,cores2024tvbench}. 
Previous works addressed this by, \eg, using LLMs to generate hard negatives \cite{momeni2023verbs}, reversing the action sequence \cite{bagad2023test}, or finer-grained objectives~\cite{wang2024paxion}. 

The closest work to ours is VideoCon~\cite{bansal2024videocon}, which finetunes a video LLM using temporally challenging hard \textit{textual} negatives.
However, our focus is different, as we explore whether generated videos can %
improve video-text alignment, complementing negative captions.

\vspace{2pt}
\noindent\textbf{Video-language alignment evaluation.} A main challenge in VLA is 
quantifying the semantic
alignment between text and video frames. 
{Early attempts used metrics}
based on the CLIPScore \cite{hessel2021clipscore,shi2022emscore,sarto2023positive}, which computes video-text alignment by measuring the similarity between video frames and their captions in the CLIP embedding space~\cite{radford2021learning}.
However, as VLMs struggle with temporal changes in captions \cite{yuksekgonul2022and, momeni2023verbs, bagad2023test, wang2024paxion}, recent approaches have started measuring video-text alignment using MLLMs for video question answering ~\cite{bansal2024videocon,lin2023revisiting,wu2024towards,li2024evaluating,wu2024vila}, such as the VQAScore in \cite{lin2024evaluating}.

In this work, we use these models to evaluate the quality of the alignment and for the new objective of evaluating how much a synthetic video aligns with its textual counterpart.

\vspace{2pt}
\noindent\textbf{Using synthetic visual data as training data.}
Recent works showed how augmenting training sets with synthetically generated images %
can improve the performance of discriminative models \cite{tian2024stablerep,peng2023synthesize,zhou2023training, he2022synthetic}. 
Diffusion models, known for their ability to generate highly realistic images and for their flexibility in dealing with different conditioning signals (text, depth, etc.), have significantly fostered this research trend~\cite{chowdhury2023apollo}. While most works focused on image recognition tasks \cite{tian2024stablerep,zhou2023training,patel2024tripletclip}, recent approaches explored more challenging tasks such as few-shot recognition \cite{he2022synthetic,samuel2024generating} or out-of-distribution detection \cite{du2024dream}. 

Our work follows a similar underlying idea and it is motivated by recent advances in text-to-video generation~\cite{yang2024cogvideox,wang2023lavie,chen2024videocrafter2,menapace2024snap}. However, we are the first to explore %
synthetic videos for improving video understanding models.

\section{Video-language alignment}
\label{sec:pf}
Video-language alignment aims to rate how well the content of a video matches a given text in natural language. Formally, let us define $\textinput$ as the given textual input in the language space $\textspace$, and $\video$ as a video in the space $\videospace$. The goal is to learn a function $\targetfunction$ parameterized by $\theta$, mapping videos and texts to their alignment scores, \ie, $\targetfunction:\videospace\times\textspace \rightarrow \left[0, 1\right]$, where $1$ means high alignment and $0$ the opposite. 

Given the fine-grained nature of language, %
this task requires video-language models with compositional and temporal order understanding and  %
recent approaches use video LLMs for this task, where an LLM is used as decoder \cite{bansal2024videocon,lin2024evaluating}.
Formally, let us define an LLM-based video-language model $\targetfunction$ via three functions: the visual encoder $\visualencoder$, the text encoder $\textencoder$, and a decoder $\textdecoder$. The two encoders map their respective inputs into a shared $d$-dimensional embedding space, %
\ie, $\visualencoder:\videospace \rightarrow \sharedspace$ and $\textencoder:\textspace \rightarrow \sharedspace$. 
The decoder maps the visual and textual inputs into a vector in the probability 
simplex $\wordssimplex$ defined over the LLM vocabulary\footnote{For simplicity, we omit the words' tokenization and we assume textual prompts and videos to be treated equally and encoded in the same space.} $\words$, \ie, $\textdecoder:\sharedspace\times\sharedspace \rightarrow \wordssimplex$. This probability vector is then used to sample the next token for the generative process. 

Within this formulation, the alignment task becomes the probability of predicting $\texttt{Yes}$ or $\texttt{No}$ as the next word after the question
$\pi_q$ = %
\texttt{Does this video entail the description [$\textinput$]}\texttt{?},
where $[t]$ is the target caption. Formally, this translates as $\targetfunction$ being:
\begin{equation}
    \label{eq:llm-prediction}
    \targetfunction(\video, \textinput) = \frac{\probyes}{\probyes+\probno} 
\end{equation}
where {$\prob = \textdecoder^{\left[\word\right]}\left( \visualencoder(\video), \textencoder(\pi_q \circ \textinput)\right)$},
with $\pi_q$ the shared question, $\circ$ string concatenation, and  $\textdecoder^{\left[\word\right]}$ the likelihood %
of the word $\word\in\words$ from the decoder's output. %

\vspace{2pt}

\noindent\textbf{VLA learning.} Usually, the parameters $\theta$ of $\targetfunction$ are updated using a dataset $D$ of $n$ video-language triplets $D=\{(\video_1,\textinput_1^{+}, \textinput_1^{-}), \cdots, (\video_n,\textinput_n^{+}, \textinput_n^{-})\}$, where $\textinput_i^{+}$ and $\textinput_i^{-}$ are the positive and negative text captions for the video $\video_i$, \ie, captions that respectively represent ($\textinput_i^{+}$) and do not represent ($\textinput_i^{-}$) the video content. Exploiting the probability distribution, output of $\textdecoder$, we can define the following objective:
\begin{equation}
    \label{eq:standard-training}
        \lossreal%
        = 
        - \sum_{i=1}^n \log \targetfunction( \video_i,\textinput^+_i) + \log \left(1-\targetfunction(\video_i,\textinput^-_i)\right).
\end{equation}
This loss function forces $\targetfunction$ to sample $\texttt{Yes}$ with a higher probability if the text represents the video and $\texttt{No}$ otherwise. 

\begin{table*}[t!]
\centering
\resizebox{0.9\textwidth}{!}{%
\begin{tabular}{C{3.3cm}|C{2.1cm}C{2.1cm}C{2.1cm}|C{2.3cm}C{2.1cm}|C{2.1cm}}
\multicolumn{1}{c|}{\textsc{Text-to-Video}} & \multicolumn{3}{b}{\textsc{Video-Language Entailment (\videocon)}} & \multicolumn{2}{r}{\textsc{Text-to-Video Retrieval}} & \cellcolor{VideoQAGreen}\textsc{Video QA} \\
\textsc{Generator} & \cellcolor{VideoConBlue}LLM & \cellcolor{VideoConBlue}Human & \cellcolor{VideoConBlue}Human-Hard & \cellcolor{RetrievalRed}SSv2-Temporal & \cellcolor{RetrievalRed}SSv2-Events & \cellcolor{VideoQAGreen}ATP-Hard \\
\midrule 
\textsc{None}* \cite{bansal2024videocon} & \textbf{88.39} & 77.16 & 74.76 & 13.00 & 10.37 & 35.46 \\
\textsc{\cogvideox} \cite{yang2024cogvideox} & 83.93 \decd{4.46} & 76.89 \decd{0.27} & 75.10 \ind{0.34} & 11.76 \decd{1.24} & 8.79 \decd{1.58} & 35.30 \decd{0.16} \\
\textsc{\lavie} \cite{wang2023lavie} & 85.26 \decd{3.13} & 76.96 \decd{0.20} & 74.63 \decd{0.13} & \textbf{14.26} \ind{1.26} & \textbf{10.80} \ind{0.43} & 34.82 \decd{0.64} \\
\textsc{\videocrafter} \cite{chen2024videocrafter2} & 85.82 \decd{2.57} & \textbf{77.33} \ind{0.17} & \textbf{75.15} \ind{0.39} & 13.80 \ind{0.80} & 10.27 \decd{0.10} & \textbf{35.79} \ind{0.33} \\
\bottomrule
\end{tabular}
}
\caption{Results of the preliminary study on using synthetic videos generated by different text-to-video models. Increases (\(\uparrow\)) and decreases (\(\downarrow\)) are measured relative to the model fine-tuned without synthetic videos (\ie, \textsc{None}). * indicates our reproduced results using the \mplugowl model checkpoint released in the original \videocon repository.}
\vspace{-5pt}
\label{tab:preliminary}
\end{table*}

\begin{table*}[t!]
\centering
\resizebox{0.9\textwidth}{!}{%
\begin{tabular}{C{3.3cm}|C{2.1cm}C{2.1cm}C{2.1cm}|C{2.3cm}C{2.1cm}|C{2.1cm}}
\multicolumn{1}{c|}{\textsc{Misalignment}} & \multicolumn{3}{b}{\textsc{Video-Language Entailment (\videocon)}} & \multicolumn{2}{r}{\textsc{Text-to-Video Retrieval}} & \cellcolor{VideoQAGreen}\textsc{Video QA}\\
& \cellcolor{VideoConBlue}LLM & \cellcolor{VideoConBlue}Human & \cellcolor{VideoConBlue}Human-Hard & \cellcolor{RetrievalRed}SSv2-Temporal & \cellcolor{RetrievalRed}SSv2-Events & \cellcolor{VideoQAGreen}ATP-Hard \\
\midrule 
\textsc{None}* \cite{bansal2024videocon} & \textbf{88.39} & 77.16 & 74.76 & 13.00 & 10.37 & 35.46 \\
\textsc{Action} & 86.10 \decd{2.29} & 77.43 \ind{0.27} & 74.83 \ind{0.07} & \textbf{15.04} \ind{2.04} & 10.66 \ind{0.29} & 36.28 \ind{0.82} \\
\textsc{Attribute} & 86.51 \decd{1.88} & 77.61 \ind{0.45} & \textbf{75.50} \ind{0.74} & 13.67 \ind{0.67} & 11.47 \ind{1.10} & 35.25 \decd{0.21} \\
\textsc{Count} & 86.10 \decd{2.29} & \textbf{77.66} \ind{0.50} & 75.27 \ind{0.51} & 14.27 \ind{1.27} & 10.97 \ind{0.60} & 36.16 \ind{0.70} \\
\rowcolor{lightgray!40}\textsc{Flip} & 85.69 \decd{2.70} & 76.04 \decd{1.12} & 73.53 \decd{1.23} & 14.94 \ind{1.94} & 10.73 \ind{0.36} & 36.06 \ind{0.60} \\
\rowcolor{lightgray!40}\textsc{Hallucination} & 85.46 \decd{2.93} & 76.55 \decd{0.61} & 74.77 \ind{0.01} & 13.89 \ind{0.89} & 10.14 \decd{0.23} & \textbf{36.37} \ind{0.91} \\
\textsc{Object} & \textbf{86.28} \decd{2.11} & 77.36 \ind{0.20} & 74.15 \decd{0.61} & 14.54 \ind{1.54} & \textbf{11.54} \ind{1.17} & 35.48 \ind{0.02} \\
\textsc{Relation} & 86.22 \decd{2.17} & 77.46 \ind{0.30} & 74.59 \decd{0.17} & 14.99 \ind{1.99} & 11.38 \ind{1.01} & 34.65 \decd{0.81} \\
\bottomrule
\end{tabular}
}
\caption{Average results of the preliminary study on using synthetic videos generated by different text-to-video models, for each type of misalignment. Increases (\(\uparrow\)) and decreases (\(\downarrow\)) are measured relative to the model fine-tuned without synthetic videos (\ie, \textsc{None}). * indicates our reproduced results using the \mplugowl model checkpoint released in the original \videocon repository. 
}
\vspace{-5pt}
\label{tab:mean_preliminary_misalignment}
\end{table*}

\begin{figure*}[t]
    \centering
    \includegraphics[width=1\linewidth]{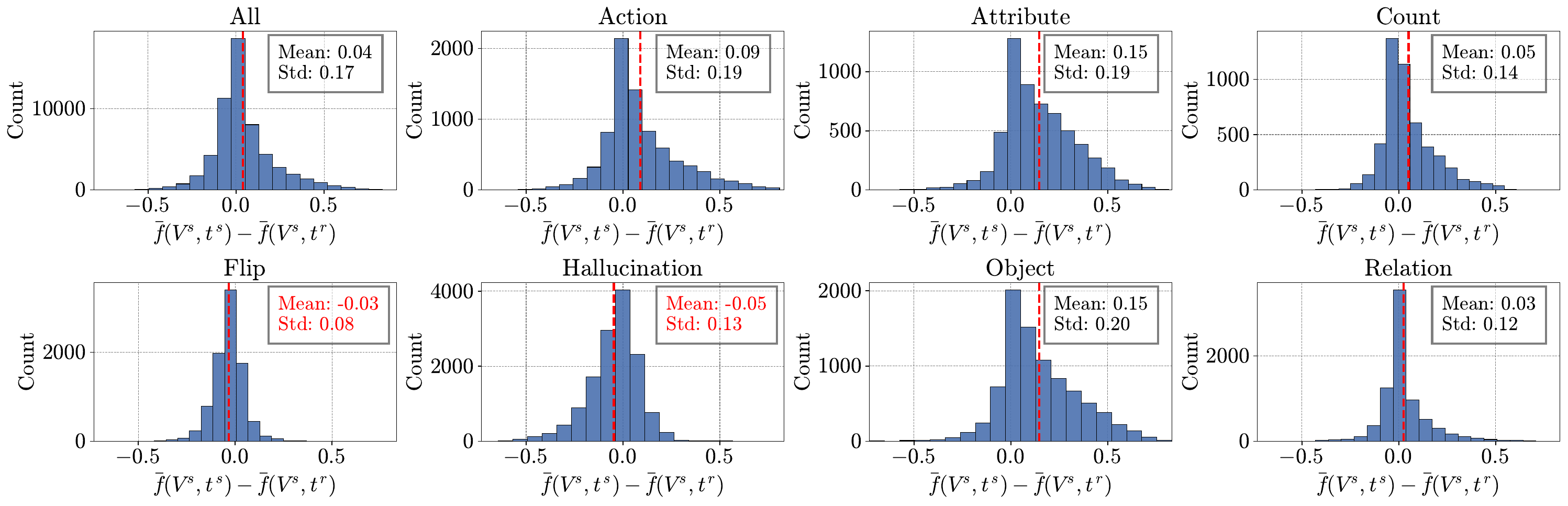}
    \caption{Distribution of the difference between $\bar{\targetfunction}(\videosynth, \textsynth)$ and $\bar{\targetfunction}(\videosynth, \textreal)$ for each misalignment type, averaged over three text-to-video generators. Misalignment types that result in negative differences (\ie, Flip and Hallucination) are highlighted in {\color{red} red}. Best viewed in color.}
    \vspace{-5pt}
    \label{fig:f_Vs_ts_minus_f_Vs_tr_distribution}
\end{figure*}

\section{Can synthetic videos help VLA?}
\label{sec:generation}
The main loss function in VLA learning, as expressed in \cref{eq:standard-training}, considers as negative only textual inputs for a given ``anchor'' video. For each positive caption $\textinput_i^+$, there is no negative video example associated with $\textinput_i^-$. 
As such, linguistic biases might be induced in the MLLM because some concepts appear only as textual negatives. Thus, we wonder: \textit{Can generated videos of negative captions help learning a VLA function?} 
To answer this question, we consider different text-to-video generator models and use them to generate synthetic videos associated to negative captions.

\vspace{2pt}

\noindent\textbf{VLA learning with generated videos.} Formally, a text-to-video generator $\generator$ maps natural language expressions in $\mathcal{T}$ and noise in the space $\noisespace$ to videos, \ie, $\generator: \textspace \times \noisespace \rightarrow \videospace$. For simplicity, we define $\textinput^r = \textinput^+$ (\ie, text %
positively associated with the \textit{real} video) and $\textinput^s = \textinput^-$ (\ie, negative text for the real video, %
positively associated with the \textit{synthetic} one). We propose to use the generator to define an objective over the dataset $D$:
\begin{equation}
    \label{eq:synth-training}
        \losssynth= %
        - \sum_{i=1}^n \log \targetfunction\left(\videosynth_i,\textsynth_i\right) %
        + \log \left(1-\targetfunction\left(\videosynth_i,\textreal_i\right)\right)
\end{equation}
where $\videosynth_i=\generator(\textsynth_i, \noise_i)$, with  $\noise_i\sim \noisespace$ being the sampled noise. The negative text $\textsynth_i$ is the input text to the generator, thus serving as the positive for the synthetic video, while the positive text $\textreal_i$ for the real video $\videoreal_i$ serves as negative for the generated video.

\vspace{2pt}
\noindent\textbf{Experimental analysis.} To better understand the potential of synthetic videos, we first conduct a preliminary experimental analysis and leverage three state-of-the-art open-source video generators, \ie, \cogvideox \cite{yang2024cogvideox}, \lavie \cite{wang2023lavie}, and \videocrafter \cite{chen2024videocrafter2}, to generate synthetic videos for each negative caption in the \videocon dataset \cite{bansal2024videocon}.
We augment the dataset with these generated videos and fine-tune a video LLM, \mplugowl \cite{ye2023mplug}, using the objective functions defined in \cref{eq:standard-training} and \cref{eq:synth-training} for real and synthetic videos, respectively. 
We measure the performance with the VLA scores estimated from \cref{eq:llm-prediction}, following the established evaluation protocol \cite{bansal2024videocon} across multiple tasks and datasets. Specifically, we consider video-language entailment on the \videocon dataset, text-to-video retrieval on SSv2-Temporal \cite{sevilla2021only} and SSv2-Events \cite{bagad2023test} datasets, and video question answering (VQA) on the ATP-Hard dataset \cite{buch2022revisiting}. The evaluation metrics include the area under the receiver operating characteristic curve (AUC ROC) on video-language entailment, mean average precision (mAP) on text-to-video retrieval, and accuracy on VQA. 

We report the results in \cref{tab:preliminary}, including baseline performance without synthetic video data (\textsc{None}). From the table, it is clear that synthetic videos harm the performance on the task closest to the training set (\ie, average drop higher than 3\% AUC on VideoCon LLM). One core reason for this drop is the distribution of the negatives being more similar to the one of the training set. Thus performance may decrease when a model sees them as positives. %
On the other hand, the results on downstream tasks suggest that synthetic videos hold promise. For instance, \videocrafter improves the result of the baseline in 4/6 settings, while \lavie boosts performance on SSv2-Temporal (\ie, +1.26 mAP). However, even with state-of-the-art video generators, not all of them guarantee improvements, and no single generator consistently outperforms the others across the tested downstream tasks. This can be seen with \cogvideox, which provides slight improvements on one of the tasks (\ie, entailment on Human-Hard) while harming the representations on the others (\eg, -1.58 mAP on SSv2-Events).

\vspace{2pt}
\noindent\textbf{Are some negative captions challenging?} The \videocon dataset \cite{bansal2024videocon} includes 
negative captions that differ from positive ones by specific types of misalignment, including modifications in actions, attributes, objects, relations, counts, event orders (flipping), and adding hallucinations. Therefore we also %
analyze whether certain types of captions are particularly challenging for the generators to produce corresponding videos. We achieve this by fine-tuning \mplugowl with synthetic videos specific to each misalignment type. The results averaged over the three video generators are reported in \cref{tab:mean_preliminary_misalignment}. 
As shown in the table, different types of misalignment have different impacts on the downstream tasks. For instance, \textsc{Action} is the misalignment that results in the largest overall improvement (\eg, +2.04 mAP on SSv2-Temporal, +0.82\% accuracy on ATP-hard), while  \textsc{Flip} and \textsc{Hallucinations} misalignments lead to some severe decrease on the VideoCon benchmarks (\eg, -1.12 and -0.61 respectively on VideoCon Human). %

We hypothesize that such a performance drop is due to the alignment quality of synthetic videos. To evaluate our hypothesis, we measure the quality of a synthetic video $\videosynth$, %
generated from a caption $\textsynth$, as a negative example for the caption $\textreal$ as $\bar{\targetfunction}(\videosynth,\textsynth) - \bar{\targetfunction}(\videosynth,\textreal)$, where $\bar{\targetfunction}(\video,\textinput)$ is computed using an ensemble of VQAScores~\cite{lin2024evaluating}, obtained by averaging the scores from three VQA models \cite{liu2024improved,instructblip, lin2024evaluating}, \ie, their average likelihood of answering $\texttt{Yes}$ to the question: \texttt{Does this figure show [$\textinput$]?} across four uniformly sampled frames from the video.
The higher the difference between the two scores, the higher the similarity of the synthetic video to its caption $\textsynth$ than its negative $\textreal$ and, intuitively, the more relevant the synthetic video for the VLA learning process. %
\cref{fig:f_Vs_ts_minus_f_Vs_tr_distribution}~shows the distribution of 
{this difference} for different types of misalignments. Notably, only \textsc{Flip} and \textsc{Hallucinations} misalignments yield mean differences that are below zero (\ie, -0.03 and -0.05, respectively), while the others are above (\eg, 0.09 \textsc{Actions},  0.15 \textsc{Attribute} and  \textsc{Object}). This 
indicates that synthetic videos corresponding to \textsc{Flip} and \textsc{Hallucinations} negative captions are not well aligned, which 
worsens the VLA learning process, as confirmed in \cref{tab:mean_preliminary_misalignment}.

\vspace{2pt}
\noindent\textbf{Finding summary.} Our preliminary analysis reveals that: \begin{enumerate*}[label=(\roman*)] \item Synthetic videos show potential for enhancing VLA, though improvements are not consistent among different generators. \item Different types of misalignment influence various downstream tasks in distinct ways. \item Synthetic videos that align closer to the positive captions of real videos rather than the negative captions result in poor training samples, which negatively impact learning. \end{enumerate*}

\begin{figure*}[t!]
    \centering
    \includegraphics[width=\linewidth]{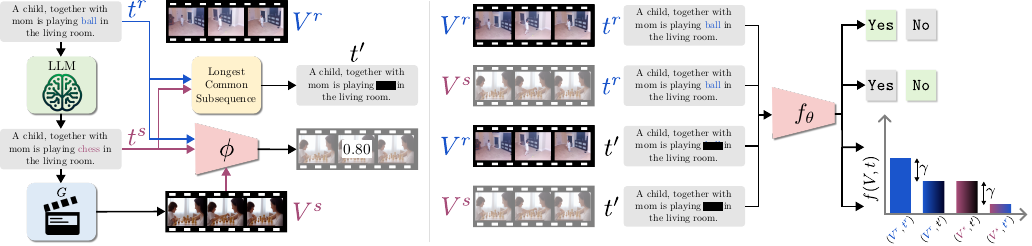}
    \caption{Overview of \textbf{\methodshort}. 
    Given a real video $\videoreal$ with its description $\textreal$ and a negative caption $\textsynth$ (generated by an LLM), we first generate a synthetic video $\videosynth$ based on $\textsynth$. We weigh the importance of each video using the scoring criterion $\phi$. We also find the shared semantic between $\textreal$ and $\textsynth$ using the longest common subsequence, obtaining $\textmasked$. We train $\targetfunction_{\theta}$ to respond with \texttt{Yes} if the input video matches its description and \texttt{No} otherwise. Additionally, we encourage the model to focus on the semantic difference between real and synthetic videos, instead of the appearance difference, using their shared semantic (\ie, $\textmasked$).}
    \vspace{-5pt}
    \label{fig:method}
\end{figure*}

\section{\methodshort}
As shown in the previous section, some generated videos are closer to real captions than their target ones (\cref{fig:f_Vs_ts_minus_f_Vs_tr_distribution}). This contradicts a key assumption of \cref{eq:synth-training}: that generated videos fully represent the content described by their input caption $\textsynth$. This often happens due to semantic inconsistency, \ie, generated videos fail to follow the semantic instruction given by the input text \cite{li2024evaluating,bansal2024videophy}. 
Such synthetic videos introduce noisy supervision signals, leading to degraded VLA performance (\cref{tab:mean_preliminary_misalignment}) \cite{lu2022lgdn}. Moreover, even semantically consistent synthetic videos may be distinguished using visual differences (\eg, artifacts~\cite{wu2024towards,polyak2025moviegencastmedia}) rather than intended semantic ones.
In this work, we propose a model-agnostic method to better use \methodfull (\methodshort), modeling them via two strategies: alignment-based weighting and semantic consistency regularization (see Fig.~\ref{fig:method}).

\vspace{2pt}
\noindent \textbf{Alignment-based weighting.}
To mitigate the impact of harmful synthetic videos and maximize the impact of valuable ones, we weigh the importance of each video based on a scoring criterion $\phi$. Given a synthetic video $\videosynth$, its corresponding caption $\textsynth$ and the real counterpart $\textreal$, $\phi$ maps %
them to a binary score in $[0,1]$ depending on their level of alignment, \ie, $\phi:\videospace\times\textspace\times\textspace\rightarrow [0,1]$. A simple choice for $\phi$ is to directly use the alignment scores given by our model $\targetfunction$. In this case, $\phi(\videosynth, \textsynth, \textreal) = \targetfunction(\videosynth,\textsynth)$. However, this might ignore the cases where, erroneously, $\targetfunction(\videosynth,\textreal)>\targetfunction(\videosynth,\textsynth)$, \ie, the generated video is closer to the real caption $\textreal$ than to the target one $\textsynth$. This phenomenon frequently happens (\ie, \cref{fig:f_Vs_ts_minus_f_Vs_tr_distribution}), due to %
\eg, wrong attribute/action binding~\cite{huang2023t2i}. For instance, if we ask the model to generate \textit{a horse watching a person running}, it may erroneously generate \textit{a person watching a horse running}, swapping the two actions. As shown in \cref{sec:generation}, this type of mistakes harms the learning of $\targetfunction$ and its capability to distinguish fine-grained details.

Thus, we define $\phi$ to account for how well the generated video $\videosynth_i$ represents $\textsynth_i$ in comparison to its real, negative, counterpart $\textreal_i$, defining the weight for a synthetic video as:
\begin{equation}
\label{eq:syn-scoring}
\weight^\phi_i =\phi(\videosynth_i, \textsynth_i, \textreal_i) = \max(0,\bar{\targetfunction}(\videosynth_i,\textsynth_i) - \bar{\targetfunction}(\videosynth_i,\textreal_i)) 
\end{equation}
where $\bar{\targetfunction}$ is %
an ensemble of VQAScores, as in \cref{sec:generation}. %
Note that the more the video is aligned with the target text \wrt its negative one, the higher its weight from Eq.~\eqref{eq:syn-scoring}.

Given this scoring criterion, we define a loss function on synthetic videos, where $\phi$ acts as a dynamic weight giving higher relevance to videos better aligned with text:
\begin{equation}
    \label{eq:synth-lambda-training}
    \begin{aligned}
        &\losssynth^\phi= %
        - \sum_{i=1}^n %
        \weight^\phi_i\cdot\left( \log \targetfunction\left(\video_i^s,\textsynth_i\right) + \log \left(1-\targetfunction\left(\videosynth_i,\textreal_i\right)\right)\right).
    \end{aligned}
\end{equation}

\vspace{2pt}
\noindent \textbf{Semantic consistency regularization.}
A positive aspect of having synthetic videos for negative textual inputs is that we can make the model focus on the semantic changes between videos rather than those in appearance. %
Suppose we are given a text $\textreal$, its negative version $\textsynth$, a real video $\videoreal$, and its generated negative version $\videosynth$. If the difference between $\textreal$ and $\textsynth$ is fine-grained, it will focus on specific %
properties of the video (\eg, action, temporal order, etc.). This implies that the two texts share most of the content \textit{but} for those fine-grained characteristics. We can thus define a text $\textmasked$, whose semantic is shared between $\textreal$ and $\textsynth$, thus not being specific to $\videoreal$ or $\videosynth$. We achieve this by finding the intersection between the two texts via longest-common subsequence~\cite{hirschberg1977algorithms}, \ie, $\textmasked = \text{LCS}(\textreal, \textsynth)$\footnote{Note that, in practice, $\textmasked$ is not implemented via token removal but via attention-level masking.}.

Note that $\textmasked$ has a specific property: given the real (synthetic) video,  $\textmasked$ is a less accurate description than the original caption $\textreal$ ($\textsynth)$, but a better one than the negative $\textsynth$ ($\textreal$). Ideally, our model should capture this relationship, modeling $\textmasked$ as semantically closer to the video than its negative caption, but farther \wrt its positive.
We can achieve this by computing a triplet loss, defined as:
\begin{equation}
    \label{eq:consistency}
    \begin{aligned}
        \lossconsis^\phi = %
         \sum_{i=1}^n \sum_{z \in \{ s, r \}} 
         &\weight^\phi_i\cdot 
        \left( \max \left( 0, \gamma + \targetfunction(\video^z_i, \textmasked_i) - \targetfunction(\video^z_i, \textinput^z_i) \right) \right. \\
        &+ \left. \max \left( 0, \gamma + \targetfunction(\video^z_i, \textinput^{\bar{z}}_i) - \targetfunction(\video^z_i, \textmasked_i) \right)
        \right).
    \end{aligned}
\end{equation}
where the margin term $\gamma$ enforces the desired separation between the alignment probabilities, and when $z=r$, $\bar{z}=s$ and vice versa. %
The first term promotes better alignment of the positive caption \wrt the generic caption $\textmasked$ and the second promotes better alignment of the latter \wrt the negative caption.  %
This encourages $\targetfunction$ to focus on the semantic differences between the two visual inputs, ignoring their differences in appearance due to the synth-to-real gap. %

\vspace{2pt}
\noindent\textbf{Full objective.} Considering all learning objectives together, we obtain the following final function:
\begin{equation}
\label{eq:full}
\loss%
= \ \lossreal %
+ \losssynth^\phi %
+ \lambdaconsis \cdot \lossconsis^\phi. %
\end{equation}
where $\lambdaconsis$ is a hyperparameter that regulates the losses.
We use Eq.~\eqref{eq:full} to learn the set $\theta$ of parameters in $\targetfunction$. %
Remarkably, our framework has only two hyperparameters, \ie, %
the margin $\gamma$ of $\lossconsis^\phi$ and the weight $\lambdaconsis$ of $\lossconsis^\phi$. %

\section{Experiments}
\label{sec:exp}

\noindent\textbf{Datasets.} For training \methodshort, we use the VideoCon dataset \cite{bansal2024videocon}, which includes temporally-challenging video-text triplets from MSR-VTT \cite{xu2016msr}, VATEX \cite{wang2019vatex}, and TEMPO \cite{hendricks2018localizing} for two tasks: \textit{Video-Language Entailment (VLE)} and \textit{Natural Language Explanation (NLE)}. In VLE, the model outputs a score of 1 if the video entails the description and 0 otherwise, while in NLE, it outputs the explanation of the differences between a video and a caption.
For each negative caption in the VideoCon VLE training set, we generate a corresponding video using three text-to-video models: \cogvideox \cite{yang2024cogvideox}, \lavie \cite{wang2023lavie}, and \videocrafter \cite{chen2024videocrafter2}.
The inference configurations for these models and examples of generated videos are in the \suppmat 

For evaluation, we use the VideoCon VLE test sets: (i) \textbf{VideoCon (LLM)}, with 27K video-text pairs from the same source datasets; (ii) \textbf{VideoCon (Human)}, with 570 pairs from ActivityNet \cite{caba2015activitynet} and human annotated negative captions; and (iii) \textbf{VideoCon (Human-Hard)}, a subset of 290 temporally challenging instances. Following \citet{bansal2024videocon}, we also evaluate our model on various downstream tasks: %
(i) text-to-video retrieval with \textbf{SSv2-Temporal} \cite{sevilla2021only}, which includes 18 action classes, each with 12 videos (in total 216 videos), requiring temporal understanding; (ii) \textbf{SSv2-Events} \cite{bagad2023test}, with 49 action classes, each with 12 videos, featuring multi-event actions; and (iii) video question answering on \textbf{ATP-Hard} \cite{buch2022revisiting}, a subset of questions of NExT-QA~\cite{xiao2021next} that require causal and temporal understanding of videos. We measure the performance using AUC for entailment, mAP for retrieval, and accuracy for VQA. Additional details on the datasets are in the \suppmat

\vspace{2pt}
\noindent\textbf{Implementation details.} We implement \methodshort on two video LLMs, \mplugowl \cite{ye2023mplug} and \videollava \cite{lin2023video}, trained on 4 NVIDIA A100 GPUs. Both models share most of the hyperparameters with \videocon \cite{bansal2024videocon} to ensure a fair comparison, and fine-tune the projection layers of the attention blocks of the LLM with low-rank adaptation (LoRA) \cite{hu2021lora}, with $r = 32$, $\alpha = 32$, and $\text{dropout} = 0.05$. 
For both models, we set $\gamma$ to 0.2, while $\lambdaconsis$ to $10^{-2}$ for \mplugowl and 1.0 for \videollava. 
Other implementation details can be found in the \suppmat

\vspace{2pt}
\noindent\textbf{Baselines.} We compare \methodshort (\mplugowl) and \methodshort (\videollava) against two sets of models. The first set includes 
{off-the-shelf VLMs} such as VideoCLIP \cite{xu2021videoclip}, ImageBind (Video-Text) \cite{girdhar2023imagebind}, \pali \cite{yarom2024you}, \mplugowl \cite{ye2023mplug}, and \videollava \cite{lin2023video}, 
{as well as models} fine-tuned for improved understanding of actions and event order, \ie, VFC \cite{momeni2023verbs} and TACT \cite{bagad2023test}. The second set consists of models trained on video-text triplets from the VideoCon dataset, namely \videocon (\mplugowl) and \videocon (\videollava) \cite{bansal2024videocon}. Additional details on the baselines are in the \suppmat

\begin{table*}[t!]
\centering
\resizebox{0.9\textwidth}{!}{%
\begin{tabular}{l|C{2.3cm}C{2.3cm}C{2.3cm}|cc|c}
 & \multicolumn{3}{b}{\textsc{Video-language Entailment (\videocon)}} & \multicolumn{2}{r}{\textsc{Text-to-Video Retrieval}} & \cellcolor{VideoQAGreen}\textsc{Video QA}\\
& \cellcolor{VideoConBlue}LLM & \cellcolor{VideoConBlue}Human & \cellcolor{VideoConBlue}Human-Hard& \cellcolor{RetrievalRed}SSv2-Temporal & \cellcolor{RetrievalRed}SSv2-Events & \cellcolor{VideoQAGreen}ATP-Hard\\
\midrule
\textsc{VideoCLIP} \cite{xu2021videoclip} & 53.2 & 47.3 & 47.5 & 9.8 & 6.4 & 23.4 \\
\textsc{ImageBind (Video-Text)} \cite{girdhar2023imagebind} & 57.1 & 65.2 & 63.0 & 10.5 & 5.5 & 25.4 \\
\textsc{TACT} \cite{bagad2023test} & - & - & - & - & 7.8 & 27.6 \\
\textsc{VFC} \cite{momeni2023verbs} & - & - & - & - & - & 31.4 \\
\textsc{\pali} \cite{yarom2024you} & 67.0 & 72.4 & 65.0 & 14.6 & 10.4 & 39.0 \\
\textsc{\mplugowl} \cite{ye2023mplug} & 57.24 & 67.02 & 64.39 & 11.08 & 6.75 & 37.96 \\
\textsc{\videollava} \cite{lin2023video} & 62.98 & 70.37 & 65.99 & 11.64 & 7.11 & 38.56 \\
\textsc{\videocon (\mplugowl)}* \cite{bansal2024videocon} & \textbf{88.39} & 77.16 & 74.76  & 13.00  & 10.37  & 35.46  \\
\textsc{\videocon (\videollava)} & 85.86 & 80.09 & 75.74 & 19.77 & 10.01 & 38.76 \\
\rowcolor{lightgray!40}\textsc{\methodshort(\mplugowl)} & 86.45 & 77.48 & 74.54 & 17.32 & \textbf{12.54} & 37.31 \\
\rowcolor{lightgray!40}\textsc{\methodshort(\videollava)} & 85.43 & \textbf{80.86} & \textbf{76.86} & \textbf{20.10} & 11.21 & \textbf{39.88} \\
\bottomrule
\end{tabular}%
}
\caption{Comparison of \methodshort with both discriminative and generative VLMs. For the \inlineColorbox{VideoConBlue}{video-language entailment} task, we report AUC-ROC, for zero-shot \inlineColorbox{RetrievalRed}{text-to-video retrieval}, we report mAP, and for \inlineColorbox{VideoQAGreen}{video question-answering}, we report accuracy.
* indicates our reproduced results using the \mplugowl model checkpoint released in the original \videocon repository. %
}
\vspace{-5pt}
\label{tab:comparison}
\end{table*}

\subsection{Comparison with state of the art}

\cref{tab:comparison} presents the results of our comparison on the VideoCon evaluation sets and the downstream tasks. Overall, our proposed method outperforms all previous baselines in five tasks out of six. For the entailment task, on the VideoCon Human dataset \methodshort (\videollava) improves its counterpart \videocon (\videollava), trained without synthetic video-caption pairs, by $0.77\%$, and achieves a $1.12\%$ improvement on its temporally challenging subset, Human-Hard. Similarly, \methodshort (\mplugowl) shows a $0.32\%$ improvement on the VideoCon Human dataset. %
As expected from \cref{sec:pf}, on the VideoCon (LLM) test set, both \methodshort (\videollava) and \methodshort (\mplugowl) underperform compared to their counterparts without synthetic videos, due to the similar distribution of negatives \wrt those present in the training set. Thus, synthetic pairs harm the performance %
in this setting. %

For text-to-video retrieval tasks, \methodshort (\mplugowl) outperforms \videocon (\mplugowl) by $4.32\%$ on SSv2-Temporal and $2.17\%$ on SSv2-Events. Similarly, \methodshort (\videollava) shows improvements of $0.33\%$ on SSv2-Temporal and $1.20\%$ on SSv2-Events compared to \videocon (\videollava). These results suggest that our model is model-agnostic and more effective at ranking similar but semantically different text descriptions than the baseline, which does not associate corresponding video data with negative captions.
Finally, for the challenging video question-answering task on the ATP-Hard dataset, models fine-tuned with only textual negatives see performance drops or minimal improvement compared to their non-finetuned version. Despite this, \methodshort (\mplugowl) improves upon \videocon (\mplugowl) by $1.85\%$, and \methodshort (\videollava) shows a $1.12\%$ improvement over \videocon (\videollava). {We show qualitative results of \methodshort in the \suppmat}

\subsection{Ablation study}

\begin{figure}[t!]
    \centering
    \includegraphics[width=0.9\linewidth]{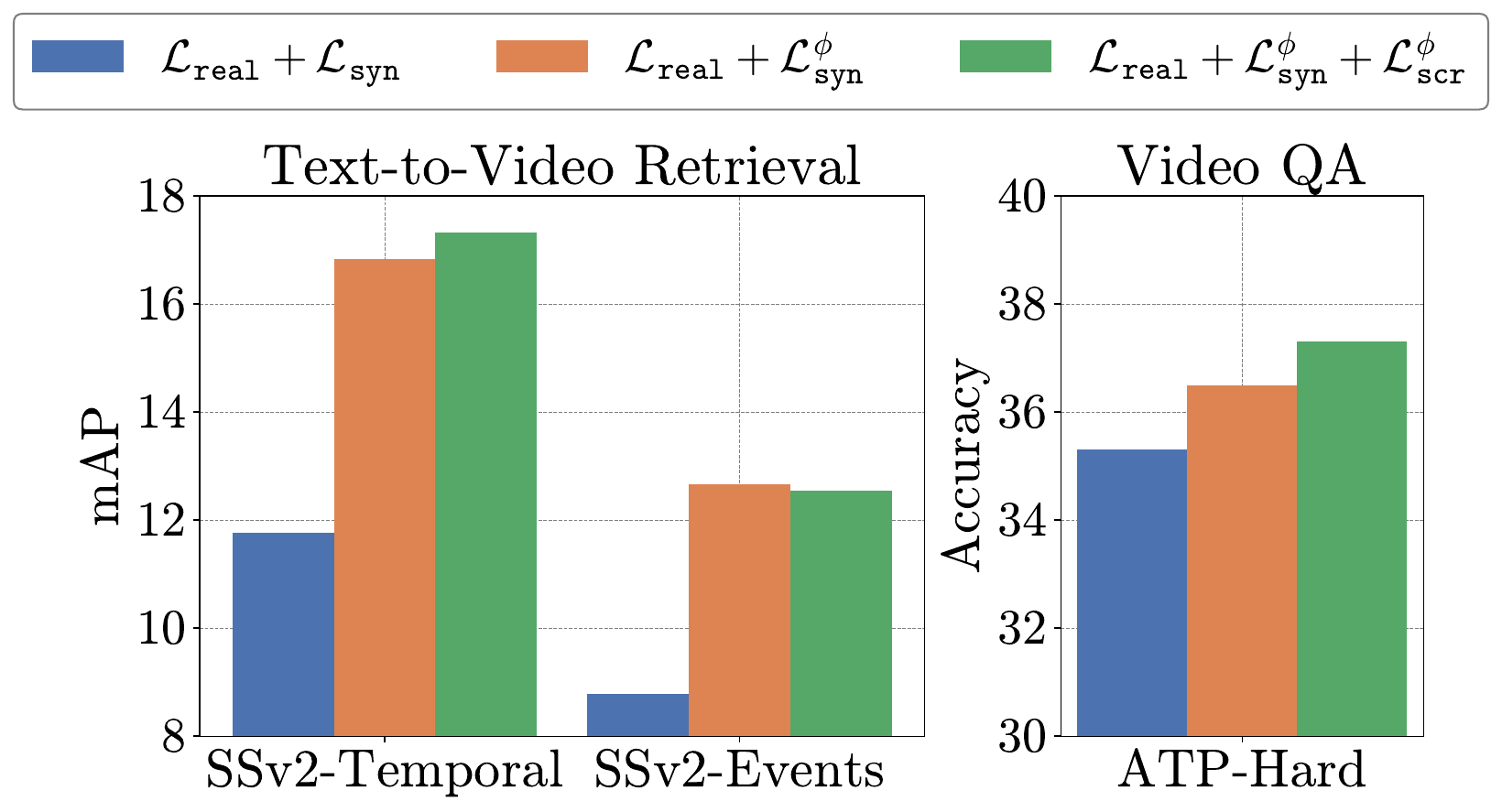}\vspace{-5pt}
    \caption{Ablation study on the proposed losses. %
    }
    \vspace{-10pt}
    \label{fig:ablation_loss}
\end{figure}

\begin{table*}[t!]
\centering
\resizebox{0.9\textwidth}{!}{%
\begin{tabular}{l|C{2.3cm}C{2.3cm}C{2.3cm}|cc|c}
& \multicolumn{3}{b}{\textsc{Video-Language Entailment (\videocon)}} & \multicolumn{2}{r}{\textsc{Text-to-Video Retrieval}} & \cellcolor{VideoQAGreen}\textsc{Video QA} \\
\textsc{Alignment-based weighting strategy} & \cellcolor{VideoConBlue}LLM & \cellcolor{VideoConBlue}Human & \cellcolor{VideoConBlue}Human-Hard & \cellcolor{RetrievalRed}SSv2-Temporal & \cellcolor{RetrievalRed}SSv2-Events & \cellcolor{VideoQAGreen}ATP-Hard \\
\midrule
1) \textsc{Fixed weighting-1.00} & 83.95 & 76.91 & \textbf{75.05} & 12.54 & 8.48 & 36.23 \\
2) $\bar{\targetfunction}(\videosynth, \textinput^s)$ & 84.87 & 76.46 & 74.12 & 13.43 & 9.40 & 35.95 \\
3) $\bar{\targetfunction}(\videosynth, \textinput^s) \cdot (1 - \bar{\targetfunction}(\videosynth, \textinput^r))$ & 85.57 & 76.79 & 74.11 & 15.52 & 10.74 & 36.06 \\
4) $\mathbbm{1}\left[ \bar{\targetfunction}(\videosynth, \textinput^s) > \bar{\targetfunction}(\videosynth, \textinput^r) \right]$
& 84.88 & 76.79 & 73.17 & 14.17 & 10.38 & 37.15 \\
\rowcolor{lightgray!40}5) $\max(0,\bar{\targetfunction}(\videosynth,\textinput^s) - \bar{\targetfunction}(\videosynth,\textinput^r))$ & \textbf{86.45} & \textbf{77.48} & 74.54 & \textbf{17.32} & \textbf{12.54} & \textbf{37.31} \\
\bottomrule
\end{tabular}
}
\caption{Results of the ablation study on the weighting strategy for the synthetic videos in the objective function.}
\vspace{-5pt}
\label{tab:ablation_dyn_weight}
\end{table*}

\begin{table}[t!]
\centering
\resizebox{\linewidth}{!}{%
\begin{tabular}{c|cccccc}
& \multicolumn{5}{c}{\textsc{Text-to-Video Generator}} \\
& \textsc{None} & \textsc{\cogvideox} & \textsc{\lavie} & \textsc{\videocrafter} & \textsc{All} \\
\midrule
\cellcolor{VideoConBlue}{LLM} & \textbf{88.39} & 86.45 & 86.45 & 86.43 & 85.82 \\
\cellcolor{VideoConBlue}{Human} & 77.16 & 77.48 & \textbf{77.51} & 77.48 & 77.15 \\
\cellcolor{VideoConBlue}{Human-Hard} & \textbf{74.76} & 74.54 & 74.73 & 74.74 & 73.79 \\
\midrule
\cellcolor{RetrievalRed}{SSv2-Temporal} & 13.00 & \textbf{17.32} & 15.98 & 15.47 & 14.06 \\
\cellcolor{RetrievalRed}{SSv2-Events} & 10.37 & \textbf{12.54} & 12.36 & 11.72 & 10.90 \\
\midrule
\cellcolor{VideoQAGreen}ATP-Hard & 35.46 & \textbf{37.31} & 36.55 & 36.50 & 36.44 \\
\bottomrule
\end{tabular}%
}
\caption{Ablation study on varying the text-to-video model.}
\vspace{-10pt}
\label{tab:ablation_generation}
\end{table}

In this section, we analyze the components of \methodshort considering the \mplugowl version. We first examine the different parts of our learning objective. We then show the benefits of alignment-based weighting over fixed weights and of different alignment-based scoring criteria. Finally, we show the effect of using different text-to-video models for generating synthetic videos. {Additional results on other designs are present in the \suppmat}

\vspace{2pt}
\noindent\textbf{Learning objectives.}  We first analyze the effectiveness of the two proposed components in our learning objective: the alignment-based loss function $\losssynth^\phi$ (\cref{eq:synth-lambda-training}) and the semantic consistency regularization $\lossconsis^\phi$ (\cref{eq:consistency}). As shown in \cref{fig:ablation_loss}, excluding $\losssynth^\phi$ leads to a drop in performance (blue vs. orange bar). Without this loss, the objective 
{is solely} the traditional language modeling loss. As a result, synthetic videos that are not aligned with their captions introduce a noisy training signal. Adding $\lossconsis^\phi$ (green bar), further boosts the performance on 2/3 datasets, suggesting that the model better captures the video semantics. As our model is trained on triplets with single-event differences \cite{bansal2024videocon}, $\lossconsis^\phi$ is less effective for SSv2-Events, where captions involve multiple events. However, current open-source video generators struggle to generate multi-event videos. 

\vspace{2pt}
\noindent\textbf{Alignment-based weighting strategy.} 
In this section, we evaluate our alignment-based weighting strategy (\ie, \cref{eq:syn-scoring}) against other alternatives, reporting the results in 
\cref{tab:ablation_dyn_weight}. As a reference, row (1) reports the results of a fixed weight (\ie, 1) for all synthetic videos. %
Assigning weights based only on alignment with the target text (\ie, $\targetfunction(\videosynth,\textinput^s)$) improves performance on retrieval (\eg, +0.92 mAP on SSv2-Events) but degrades performance on others (\eg, on ATP-Hard, -0.28\%), as it overlooks cases where synthetic videos align more with real captions. In row (3), we multiply the synthetic scores by the inverse similarity with the real counterpart (\ie, $(1 - \targetfunction(\videosynth,\textinput^r))$). Introducing the real captions into the score improves the results in various settings, especially on retrieval, achieving +2.98 mAP on SSv2-Temporal, and +2.26 mAP on SSv2-Events. As an alternative, row (4) considers weighing all synthetic videos as 1 if they are closer to their target caption than the real one. This strategy shows a general degradation w.r.t. the previous, except for ATP-Hard (+1.9\%). This denotes that a soft-weighting scheme is still more effective as it accounts for different levels of semantic fidelity across videos. Our proposed strategy (row (5)) combines the advantages of the two, enforcing that synthetic videos are truly negative examples, \ie, being more similar to their caption than the original one of the real videos.  This strategy obtains the highest %
results in almost all settings. %
{For \videollava, we use (3) as it performs slightly better.}

\vspace{2pt}
\noindent\textbf{Text-to-video generators.}  
We analyze this aspect by comparing three text-to-video generators when used with our method: \cogvideox \cite{yang2024cogvideox}, \lavie \cite{wang2023lavie}, and \videocrafter \cite{chen2024videocrafter2}. As shown in \cref{tab:ablation_generation}, \methodshort (\mplugowl) fine-tuned on videos generated by \cogvideox outperforms the other alternatives across all downstream tasks and achieves comparable results on the video-language entailment task. While it can be challenging to determine \textit{a-priori} the optimal generator for a downstream task, one possibility could be to generate videos from multiple generators and let the model filter them. Using all generated videos performs better than using none on the downstream tasks (\eg, +1.06\% on SSv2-Temporal), but underperforms CogVideoX (\eg, -3.26\% on SSv2-Temporal). 
This is likely due to the high synth-to-real video ratio, introducing a significant domain shift that requires careful handling. Nevertheless, we expect that the better the text-to-video models released, the more beneficial they will be for \methodshort.

\section{Conclusion}
\label{sec:conclusion}

In this work, we explored whether videos generated by text-to-video models can help learning a better video-language alignment (VLA) model. %
Our initial analysis shows that synthetic videos can boost performance on certain downstream tasks, but harm others. We attribute this to (i) semantic inconsistency, as synthetic videos may not follow the input text, and (ii) appearance bias, where the model focuses on visual differences in the videos rather than semantic differences. To address these limitations, we introduced \methodshort, the first VLA method exploiting synthetic videos. \methodshort includes an alignment-based sample weighting strategy to mitigate noisy video generations and a semantic consistency regularization to make the model focus on semantic, rather than visual, differences. \methodshort outperforms baselines that do not use synthetic videos across different video LLMs on five out of six tasks, demonstrating its potential to improve VLA across diverse models.

\noindent\textbf{Acknowledgments.}
This work was sponsored by EU ISFP PRECRISIS (ISFP2022-TFI-AG-PROTECT-02-101100539), PNRR ICSC National Research Centre for HPC, Big Data and Quantum Computing (CN00000013), Ministero delle Imprese e del Made in Italy (IPCEI Cloud DM 27 giugno 2022 – IPCEI-CL-0000007), the FAIR - Future AI Research (PE00000013), funded by NextGeneration EU, and EU PATTERN (Project No. 101159751). We acknowledge ISCRA for awarding this project access to the LEONARDO supercomputer, owned by the EuroHPC Joint Undertaking, hosted by CINECA (Italy).

{
    \small
    \bibliographystyle{ieeenat_fullname}
    \bibliography{main}
}

\clearpage
\renewcommand{\thesection}{\AlphAlph{\value{section}-8}}

\maketitlesupplementary

\noindent In this supplementary material, we provide additional details about our implementation (\cref{sec:supp_implementation}), the training and evaluation datasets (\cref{sec:supp_datasets}), performance metrics (\cref{sec:supp_metrics}), and baseline models (\cref{sec:supp_baselines}). We also provide further analyses of design choices in \cref{sec:supp_ablation}, along with the inference configuration of text-to-video models and examples of generated videos in \cref{sec:supp_inference}. Finally, we describe the limitations of our method in \cref{sec:supp_limitations} and present qualitative results on evaluation datasets in \cref{sec:supp_qualitatives}.

\section{Implementation details}
\label{sec:supp_implementation}

We implement \methodshort on two video large language models (LLMs): \mplugowl \cite{ye2023mplug} and \videollava \cite{lin2023video}, trained on 4 NVIDIA A100 GPUs with 64GB memory each. We train the same layers %
as in \videocon \cite{bansal2024videocon} to ensure a fair comparison. Specifically, we fine-tune the projection layers of the attention blocks of the LLM using low-rank adaptation (LoRA) \cite{hu2021lora} with parameters $r = 32$, $\alpha = 32$, and $\text{dropout} = 0.05$. We train the model for 1 epoch for both \mplugowl and \videollava. Due to memory constraints, we adjust the batch sizes for each model: \mplugowl uses a batch size of 8, and \videollava uses a batch size of 4. 
Both models are trained using the Adam optimizer \cite{kingma2014adam}, with a linear warmup of 200 steps, a cosine annealing schedule, and learning rates of $10^{-4}$ for \mplugowl and $5 \times 10^{-5}$ for \videollava. 
We empirically set the margin term $\gamma$ to 0.2 for both models and %
the consistency weight $\lambdaconsis$ to $10^{-2}$ for \mplugowl and 1.0 for \videollava. 

\section{Details about datasets}
\label{sec:supp_datasets}

For training \methodshort, we use the VideoCon dataset \cite{bansal2024videocon}, which includes temporally-challenging video-text triplets from MSR-VTT \cite{xu2016msr}, VATEX \cite{wang2019vatex}, and TEMPO \cite{hendricks2018localizing} for two tasks: \textit{Video-Language Entailment (VLE)} and \textit{Natural Language Explanation (NLE)}. In VLE, the model outputs a score of 1 if the video entails the description and 0 otherwise. In NLE, the model outputs an explanation of the difference between a video and a caption.

For training, we use the same training split of \videocon \cite{bansal2024videocon}, containing about 108K video-text triplets for both tasks. 
The VideoCon dataset includes negative captions generated to represent seven types of semantically plausible misalignments: \textit{object, action, attribute, counting, relation, hallucination}, and \textit{event order flip}. For each negative caption, we generate a corresponding video using three open-source text-to-video generation models: \cogvideox \cite{yang2024cogvideox}, \lavie \cite{wang2023lavie}, and \videocrafter \cite{chen2024videocrafter2}, resulting in a total of 173,337 generated videos.

For evaluation, we use the VLE test sets of VideoCon, which include i)
\textbf{VideoCon (LLM):} 27K video-text pairs from the same datasets used for training, with negative captions generated by an LLM;
ii) \textbf{VideoCon (Human):} 570 pairs from ActivityNet \cite{caba2015activitynet}, where negative captions are manually annotated;
iii) \textbf{VideoCon (Human-Hard):} a subset of VideoCon (Human) with 290 pairs labeled as temporally challenging (\ie, each video frame does not entail the caption) by an image-text alignment model \cite{yarom2024you}.

Following \cite{bansal2024videocon}, we also evaluate our model on downstream tasks that require temporal understanding. Specifically, we test on: i) text-to-video retrieval using \textbf{SSv2-Temporal}~\cite{sevilla2021only} and \textbf{SSv2-Events}~\cite{bagad2023test}; and ii) on video question answering using \textbf{ATP-Hard}~\cite{buch2022revisiting}.
The SSv2-Temporal dataset includes 18 action classes, each with 12 matching videos (in total 216 videos), featuring actions %
such as \textit{Moving [something] and [something] away from each other}.
The SSv2-Events dataset contains 49 action classes with 12 videos per class.
SSv2-Events focuses on templates involving multiple verbs, which indicates various events within a video, such as \textit{Pouring [something] into [something] until it overflows}.
Finally, ATP-Hard \cite{buch2022revisiting} is a subset of questions of the NExT-QA benchmark~\cite{xiao2021next} that require causal and temporal understanding of videos where, for each question, there are 5 possible answers.

\section{Details about performance metrics}
\label{sec:supp_metrics}
We follow the evaluation protocol established by \citet{bansal2024videocon} and measure the video-language entailment performance using the area under the receiver operating characteristic curve (AUC ROC) on the VLE test sets of VideoCon. 
For text-to-video retrieval, we compute video-language alignment scores between each action class and the candidate videos, rank the videos, and report the mean Average Precision (mAP).
For video question answering, we use statements generated by \citet{bansal2024videocon} via the PaLM-2 API \cite{anil2023palm}. We compute the alignment scores of these statements with the input video, select the highest-ranking statement, and report the accuracy.

\section{Details about baselines}
\label{sec:supp_baselines}

We implement both \textsc{\videocon} baselines and our \methodshort with two video large language models: 
\mplugowl \cite{ye2023mplug} and \videollava \cite{lin2023video}. \textit{\mplugowl} employs a CLIP ViT-L/14 \cite{dosovitskiy2020image} visual encoder to extract features from 32 uniformly sampled video frames. These features are processed by a visual abstractor module, which includes additional temporal query tokens for temporal modeling, to compress the visual information into a fixed number of learnable tokens. The resulting tokens are combined with tokenized textual queries and provided as input into LLaMA-7B \cite{touvron2023llama}, serving as the large language model. \textit{\videollava} employs LanguageBind encoders \cite{zhu2023languagebind}, initialized from CLIP ViT-L/14 \cite{dosovitskiy2020image}, to map 8 uniformly sampled video frames into the textual feature space of LanguageBind \cite{zhu2023languagebind}. A 2-layer fully connected network processes these features, which are then combined with tokenized textual queries and fed as input to Vicuna-7B v1.5 \cite{chiang2023vicuna}, serving as the large language model.

For the baseline \textsc{\videocon (\mplugowl)}, we report results in the main manuscript using the checkpoint from the latest version of the official \videocon repository, which includes the corrected LoRA $\alpha$ parameter introduced in commit \texttt{9a69520}.

For the baseline \textsc{\videocon (\videollava)}, we report results obtained by fine-tuning the \videollava model on the \videocon dataset using the same trained layers and hyperparameters as in \videocon \cite{bansal2024videocon}. Specifically, we train the model for 2 epochs, fine-tuning the projection layers of the LLM's attention blocks using LoRA with parameters $r = 32$, $\alpha = 32$, and $\text{dropout} = 0.05$. We use a batch size of 16, the Adam \cite{kingma2014adam} optimizer, a linear warmup of 200 steps, a cosine annealing learning rate schedule, and a learning rate of $10^{-4}$.

Finally, apart from the off-the-shelf versions of \mplugowl and \videollava, for which we performed evaluation on the evaluation sets, the results of all other baselines (\ie, VideoCLIP \cite{xu2021videoclip}, ImageBind (Video-Text) \cite{girdhar2023imagebind}, \pali \cite{yarom2024you}, VFC \cite{momeni2023verbs}, and TACT \cite{bagad2023test}) are taken from the \videocon paper \cite{bansal2024videocon}.

\section{Additional analysis}
\label{sec:supp_ablation}

\begin{table}[t!]
\centering
\resizebox{\columnwidth}{!}{%
\begin{tabular}{l|C{2.1cm}C{2.1cm}C{2.1cm}}
 & \multicolumn{3}{c}{\cellcolor{VideoConBlue}\textsc{Video-language Entailment (\videocon)}} \\
 & \cellcolor{VideoConBlue}LLM & \cellcolor{VideoConBlue}Human & \cellcolor{VideoConBlue}Human-Hard \\
\midrule
\textsc{\instructblip} \cite{lin2024evaluating} & 64.97 & 73.37 & 66.87 \\
\textsc{\llava} \cite{lin2024evaluating} & 64.54 & 69.84 & 63.67 \\
\textsc{\clipflantfive} \cite{lin2024evaluating} & 65.59 & 74.41 & 67.92 \\
\textsc{VQAScores \cite{lin2024evaluating}} & 66.60 & 74.60 & 67.85 \\
\rowcolor{lightgray!40}\textsc{\methodshort(\mplugowl)} & \textbf{86.45} & 77.48 & 74.54 \\
\rowcolor{lightgray!40}\textsc{\methodshort(\videollava)} & 85.43 & \textbf{80.86} & \textbf{76.86} \\
\bottomrule
\end{tabular}%
}
\caption{Comparison of \methodshort with image-text alignment models \cite{liu2024improved, instructblip, lin2024evaluating}.
}
\label{tab:ablation_vqascores}
\end{table}

\begin{table}[t!]
\centering
\resizebox{0.9\columnwidth}{!}{%
\begin{tabular}{c|C{2.1cm}C{2.1cm}C{2.1cm}}
& \multicolumn{3}{c}{\cellcolor{VideoConBlue}\textsc{Video-Language Entailment (\videocon)}} \\
\textsc{Margin} & \cellcolor{VideoConBlue}LLM & \cellcolor{VideoConBlue}Human & \cellcolor{VideoConBlue}Human-Hard \\
\midrule 
\rowcolor{lightgray!40}0.2 & \textbf{86.14} & 77.25 & \textbf{73.95} \\
0.4 & 86.04 & \textbf{77.39} & 73.71 \\
0.6 & 86.03 & 77.17 & 73.72 \\
0.8 & 85.99 & 76.78 & 73.39 \\
\bottomrule
\end{tabular}
}
\caption{Results of varying the margin term used for semantic consistency regularization.}
\label{tab:ablation_triplet_margin}
\end{table}

\begin{table}[t!]
\centering
\resizebox{0.9\columnwidth}{!}{%
\begin{tabular}{c|C{2.1cm}C{2.1cm}C{2.1cm}}
& \multicolumn{3}{c}{\cellcolor{VideoConBlue}\textsc{Video-Language Entailment (\videocon)}} \\
\textsc{Weight} & \cellcolor{VideoConBlue}LLM & \cellcolor{VideoConBlue}Human & \cellcolor{VideoConBlue}Human-Hard \\
\midrule 
$10^{-3}$ & \textbf{86.45} & \textbf{77.48} & 74.53 \\
\rowcolor{lightgray!40} $10^{-2}$ & \textbf{86.45} & \textbf{77.48} & \textbf{74.54} \\
$10^{-1}$ & 86.40 & 77.34 & 74.29 \\
1.0 & 86.14 & 77.25 & 73.95 \\
\bottomrule
\end{tabular}%
}
\caption{Results of varying the weight term used for semantic consistency regularization.}
\label{tab:ablation_lambdaconsis}
\end{table}

\begin{table}[t!]
\centering
\resizebox{\linewidth}{!}{%
\begin{tabular}{c|cccc}
& \multicolumn{4}{c}{\textsc{Misalignment Type}} \\
& \textsc{Action} & \textsc{Action (\methodshort)} & \textsc{Hall.} & \textsc{Hall. (\methodshort)} \\
\midrule
\cellcolor{VideoConBlue}{LLM} & 86.10 & 85.80 \decd{0.30} & 85.46 & \textbf{86.45} \ind{0.99}\\
\cellcolor{VideoConBlue}{Human} & 77.43 & \textbf{77.86} \ind{0.43} & 76.55 & 77.03 \ind{0.48}\\
\cellcolor{VideoConBlue}{Human-Hard} & 74.83 & \textbf{74.96} \ind{0.13} & 74.77 & 74.74 \decd{0.03}\\
\midrule
\cellcolor{RetrievalRed}{SSv2-Temporal} & 15.04 & \textbf{15.41} \ind{0.37} & 13.89 & 14.89 \ind{1.00}\\
\cellcolor{RetrievalRed}{SSv2-Events} & 10.66 & \textbf{11.66} \ind{1.00} & 10.14 & 11.00 \ind{0.86}\\
\midrule
\cellcolor{VideoQAGreen}ATP-Hard & 36.28 & \textbf{37.62} \ind{1.34} & 36.37 & 36.42 \ind{0.05}\\
\bottomrule
\end{tabular}%
}
\caption{Average results of applying \methodshort to specific misalignment types across different text-to-video models. Increases (\(\uparrow\)) and decreases (\(\downarrow\)) are measured relative to the model ``blindly'' fine-tuned with synthetic videos of the same misalignment type.}
\label{tab:synvita_misalignment}
\end{table}

\begin{table*}[t!]
\centering
\resizebox{\linewidth}{!}{%
\begin{tabular}{@{}lccccccc@{}}
\toprule
\textbf{Model} & 
\textbf{Resolution (W$\times$H)} & 
\textbf{Length (frames)} & 
\textbf{FPS (frames/s)} & 
\textbf{Guidance Scale} & 
\textbf{Sampling Steps} & 
\textbf{Noise Scheduler} &
\textbf{Generation Time (s)} \\ 
\midrule
\textsc{\cogvideox} \cite{yang2024cogvideox}    & 720$\times$480 & 49 & 8 & 6.0  & 50 & DDIM \cite{song2020denoising} & $\sim$75 \\
\textsc{\lavie} \cite{wang2023lavie}        & 512$\times$320 & 32 & 8 & 7.5  & 50 & DDPM \cite{ho2020denoising} & $\sim$20 \\
\textsc{\videocrafter} \cite{chen2024videocrafter2} & 512$\times$320 & 32 & 8 & 12.0 & 50 & DDIM \cite{song2020denoising} & $\sim$109 \\
\bottomrule
\end{tabular}%
}
\caption{Inference configurations for text-to-video generators.}
\label{tab:t2v_inference}
\end{table*}

In this section, we analyze the models used to estimate alignment scores for our alignment-based weighting strategy, the hyperparameters of our model, \ie, the margin and the weight of the semantic consistency regularization, and \methodshort's application to specific types of misalignment.

\noindent\textbf{Ensemble for image-text alignment models.} 
Our alignment-based weighting strategy relies on alignment scores between synthetic videos and captions. We thus analyze the performance of the state-of-the-art approach \cite{lin2024evaluating} for computing these scores on the video-language entailment task, which is the closest task to image-text alignment. 
Specifically, we evaluate three multimodal LLMs: \textsc{\llava} \cite{liu2024improved}, \textsc{\instructblip} \cite{instructblip}, and \textsc{\clipflantfive} \cite{lin2024evaluating}. In addition to using them individually, we include an ensemble version referred to as \textsc{VQAScores} \cite{lin2024evaluating}, which averages the scores of the image-text alignment models. 

As shown in Table \ref{tab:ablation_vqascores}, the ensemble \textsc{VQAScores} generally outperforms individual models, as videos that consistently receive high alignment scores across all models are more likely to be consistent with their textual description. 
In \methodshort, we use this ensemble to evaluate the quality of the alignment, as this offers more precise scores than individual models.
For reference, we also report the results obtained by \methodshort (\mplugowl) and \methodshort (\videollava). These results highlight a performance gap between models fine-tuned on the video-language entailment task and off-the-shelf multimodal LLMs. Specifically, \methodshort (\videollava), which relies on a smaller LLM (7B, compared to the 13B for \textsc{\llava}, and 11B for \textsc{\instructblip} and \textsc{\clipflantfive}), improves the ensemble's performance by 18.83\%, 6.26\%, and 9.01\% on the VideoCon evaluation datasets. This supports the finding that off-the-shelf multimodal LLMs often lack robustness to fine-grained caption manipulations \cite{li2024evaluating}. Nevertheless, we can use their prior knowledge to estimate the semantic consistency of the generated videos.

\noindent\textbf{Margin term $\gamma$ of $\lossconsis^\phi$.} The margin term $\gamma$ of the semantic consistency regularization loss controls the desired separation between alignment probabilities. We analyze its effect by fixing the loss weight to 1 and present the results in \cref{tab:ablation_triplet_margin}. As can be seen, setting $\gamma$ to 0.2 for \mplugowl achieves the best performance on two out of three video-language entailment datasets.

\noindent\textbf{Weight $\lambdaconsis$ of $\lossconsis^\phi$.} The weight $\lambdaconsis$ of the semantic consistency regularization regulates its contribution to the overall objective. We analyze the effect of varying this hyperparameter and report the results in \cref{tab:ablation_lambdaconsis}. Setting the value to \(10^{-2}\) for \mplugowl achieves the highest performance across all video-language entailment datasets.

\noindent\textbf{\methodshort on specific types of misalignment.} \cref{tab:mean_preliminary_misalignment} shows how different types of misalignment affect downstream tasks. A natural question is how \methodshort performs when applied to a specific misalignment. To explore this, we select one misalignment with a positive mean (\ie, \textsc{action}) and one with a negative mean (\ie, \textsc{hallucination}) from \cref{fig:f_Vs_ts_minus_f_Vs_tr_distribution}. We then apply \methodshort to videos of these categories, generated by three video generators, and report the average results in \cref{tab:synvita_misalignment}. The results show that \methodshort improves performance on 5 out of 6 tasks (\eg, +1\% and +0.86\% on SSv2-Events) compared to the model ``blindly'' fine-tuned on the same synthetic videos.

\section{Analysis of text-to-video generators}
\label{sec:supp_inference}

\begin{figure*}[t!]
    \centering
    \includegraphics[width=\linewidth]{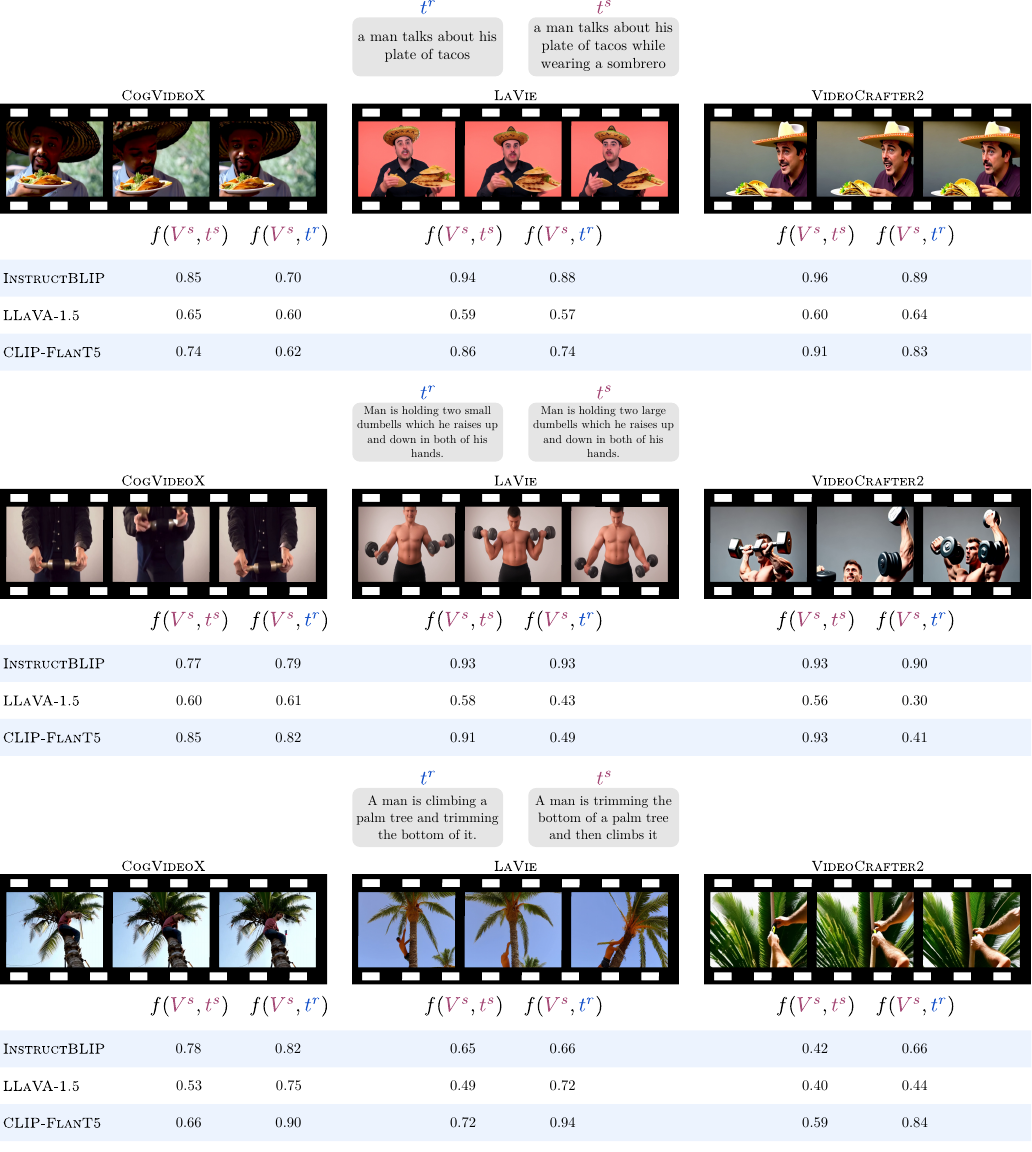}
    \caption{Examples of videos generated by three text-to-video models (\ie, \cogvideox, \lavie, and \videocrafter) from LLM-generated negative captions, along with alignment scores assigned by different image-text alignment methods (\ie, \instructblip, \llava, and \clipflantfive). For each synthetic video $\videosynth$ and alignment model, we show its alignment with the corresponding caption $\textsynth$, denoted as $\targetfunction(\videosynth, \textsynth)$, and with the real caption $\textreal$, denoted as $\targetfunction(\videosynth, \textreal)$.}
    \label{fig:video_generation}
\end{figure*}

For each negative caption in the \videocon dataset \cite{bansal2024videocon}, we generate a corresponding video using three open-source text-to-video generation models: \cogvideox \cite{yang2024cogvideox}, \lavie \cite{wang2023lavie}, and \videocrafter \cite{chen2024videocrafter2}, resulting in a total of 173,337 generated videos. The inference configurations used for these models are reported in \cref{tab:t2v_inference}. For each model, we use the default configuration available at the time of cloning the respective GitHub repository, with the following exceptions: for \lavie and \videocrafter, we set the number of frames to 32 and the frames per second to 8 to ensure that the generated videos have a duration of 4 seconds (the longest we can obtain with the available models). As shown in the table, for \cogvideox, we generate longer videos (approximately 6.125 seconds) with a higher resolution ($720\times480$ pixels) compared to the 4-second videos and $512\times320$ pixel resolution of \lavie and \videocrafter. Using NVIDIA A100 GPUs, the average generation times per video are:
(1) \cogvideox: 1.51s per step, $\sim$75s total.
(2) \lavie: 0.4s per step, $\sim$20s total.
(3) \videocrafter: 2.18s per step, $\sim$109s total.

\cref{fig:video_generation} presents examples of videos generated from LLM-generated negative captions, along with alignment scores assigned by image-text alignment methods (\ie, \instructblip \cite{instructblip}, \llava \cite{liu2024improved}, and \clipflantfive \cite{lin2024evaluating}). Specifically, we show the alignment of each synthetic video $\videosynth$ with its corresponding caption $\textsynth$, denoted as $\targetfunction(\videosynth, \textsynth)$, and with its real counterpart $\textreal$, denoted as $\targetfunction(\videosynth, \textreal)$. For the caption \textit{A man talks about a plate of tacos while wearing a sombrero}, all three models generate semantically consistent videos, resulting in higher alignment scores for the input caption than the real counterpart. In contrast, for the caption \textit{A man is holding two large dumbbells which he raises up and down in both hands}, \cogvideox fails to depict the size of the dumbbells. In this case, the alignment between the synthetic video and the corresponding caption is lower than that of the real counterpart for two out of three models.
Finally, for the caption \textit{A man is trimming the bottom of a palm tree and then climbs it}, none of the generated videos achieve higher alignment scores with the input caption than with the real caption. This is likely due to the nonsensical nature of the caption generated by the LLM. As this caption belongs to the \textit{event order flip} type of misalignment, if many captions of this type are similarly affected by this issue, it could explain the negative mean alignment difference observed in 
\cref{fig:f_Vs_ts_minus_f_Vs_tr_distribution}
for this category of synthetic videos. 
Such negative mean may also result from artifacts introduced by text-to-video models. This is particularly evident in the video generated by \videocrafter, where the model only includes the top of a palm tree to adhere to the input prompt. 

\section{Limitations}
\label{sec:supp_limitations}
Our method depends on the capability of text-to-video generators, which are constrained to produce short-duration videos, often shorter than real ones. This may contribute to a larger syn-real shift and lead to less temporally challenging synthetic videos, limiting the learning strength of our method. While our work pioneers this research direction, further benefits will come from future advances in high-quality video generation and alignment evaluation.

\section{Qualitative results of \methodshort}
\label{sec:supp_qualitatives}

\begin{figure*}[t!]
    \centering
    \includegraphics[width=0.93\linewidth]{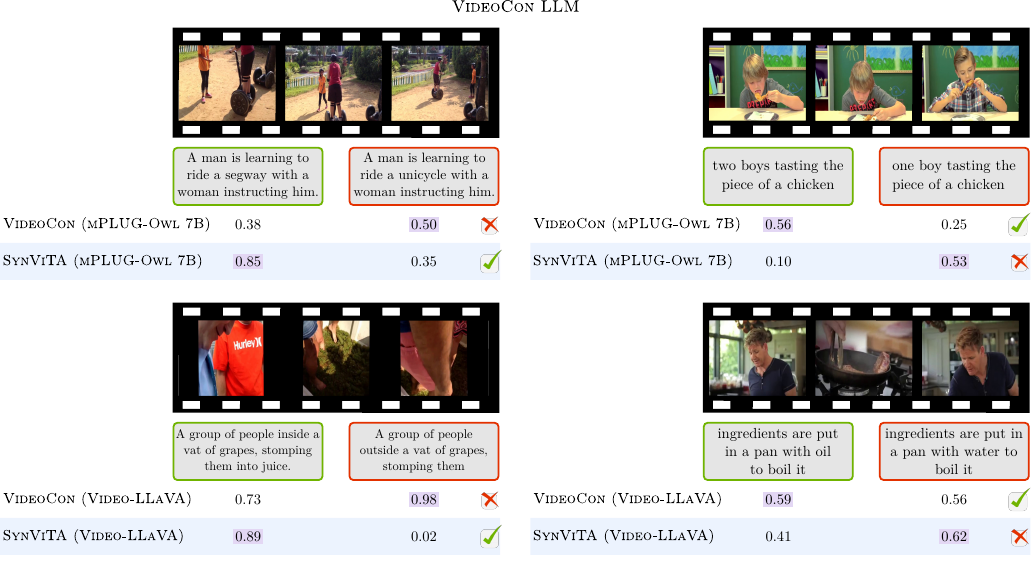}
    \caption{Examples of video-language alignment scores assigned by \methodshort (\mplugowl) and \methodshort (\videollava), compared to baselines trained without synthetic videos, for the video-language entailment task on \videocon LLM. Captions marked with {\color{forestgreen(web)}green} borders correctly match the input video, while those marked with {\color{red}red} borders do not. The models’ highest predicted scores are highlighted in {\color{violet}violet}. If the top prediction corresponds to the caption that correctly describes the video, the row is marked with a checkmark; otherwise, it is marked with a cross.}
    \label{fig:videocon_llm}
\end{figure*}

\begin{figure*}[t!]
    \centering
    \includegraphics[width=0.93\linewidth]{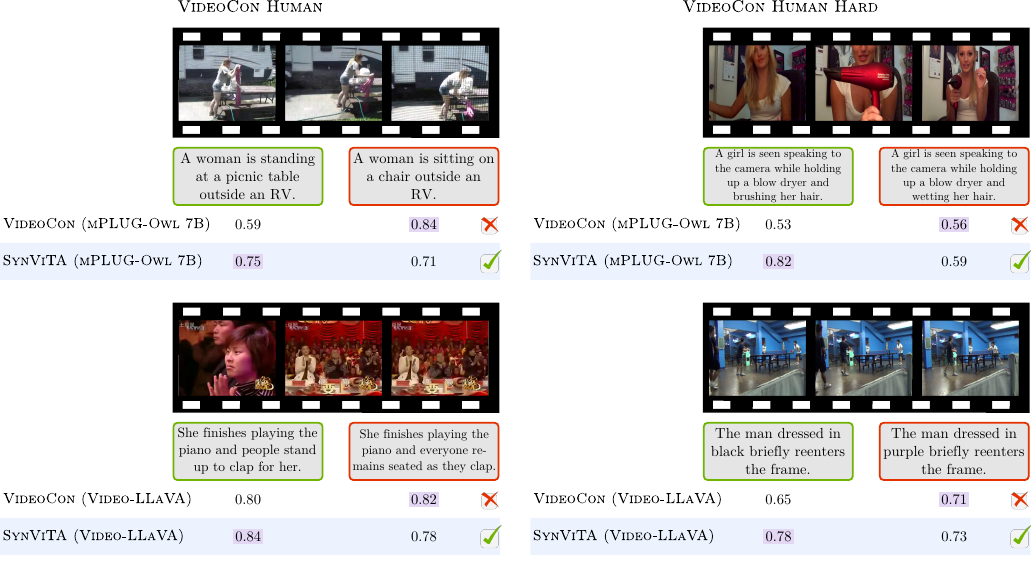}
    \caption{Examples of video-language alignment scores assigned by \methodshort (\mplugowl) and \methodshort (\videollava), compared to baselines trained without synthetic videos, for the video-language entailment task on \videocon Human and Human Hard. Captions marked with {\color{forestgreen(web)}green} borders correctly match the input video, while those marked with {\color{red}red} borders do not. The models’ highest predicted scores are highlighted in {\color{violet}violet}. If the top prediction corresponds to the caption that correctly describes the video, the row is marked with a checkmark; otherwise, it is marked with a cross.}
    \label{fig:videocon_human}
\end{figure*}

\begin{figure*}[t!]
    \centering
    \includegraphics[width=0.95\linewidth]{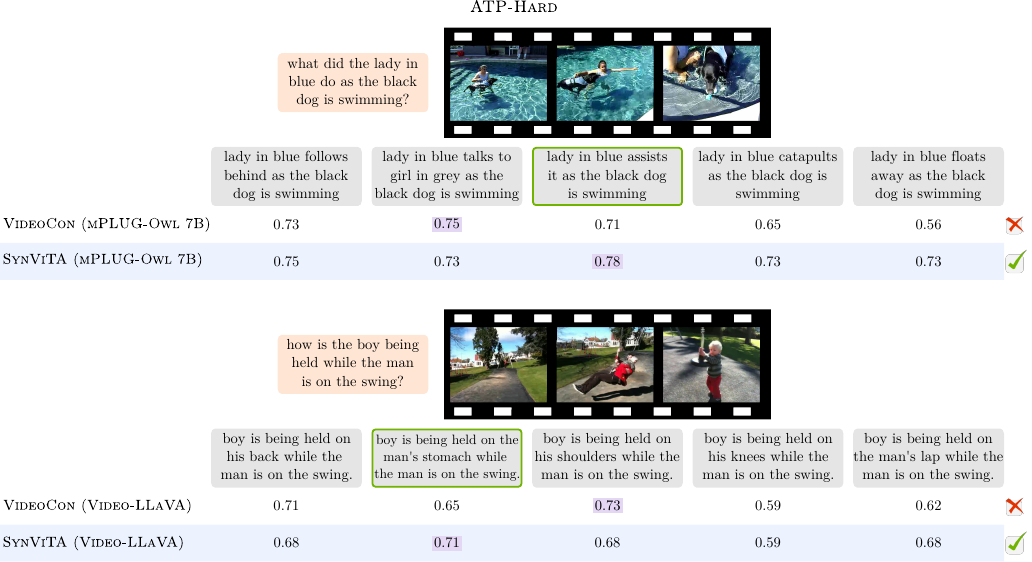}
    \caption{Examples of video-language alignment scores assigned by \methodshort (\mplugowl) and \methodshort (\videollava), compared to baselines trained without synthetic videos, on the video question answering task for ATP-Hard. Captions marked with {\color{forestgreen(web)}green} borders are the correct answers. The models’ highest predicted scores are highlighted in {\color{violet}violet}. If the top prediction corresponds to the correct answer, the row is marked with a checkmark; otherwise, it is marked with a cross.}
    \label{fig:atphard}
\end{figure*}

\begin{figure*}[t!]
    \centering
    \includegraphics[width=\linewidth]{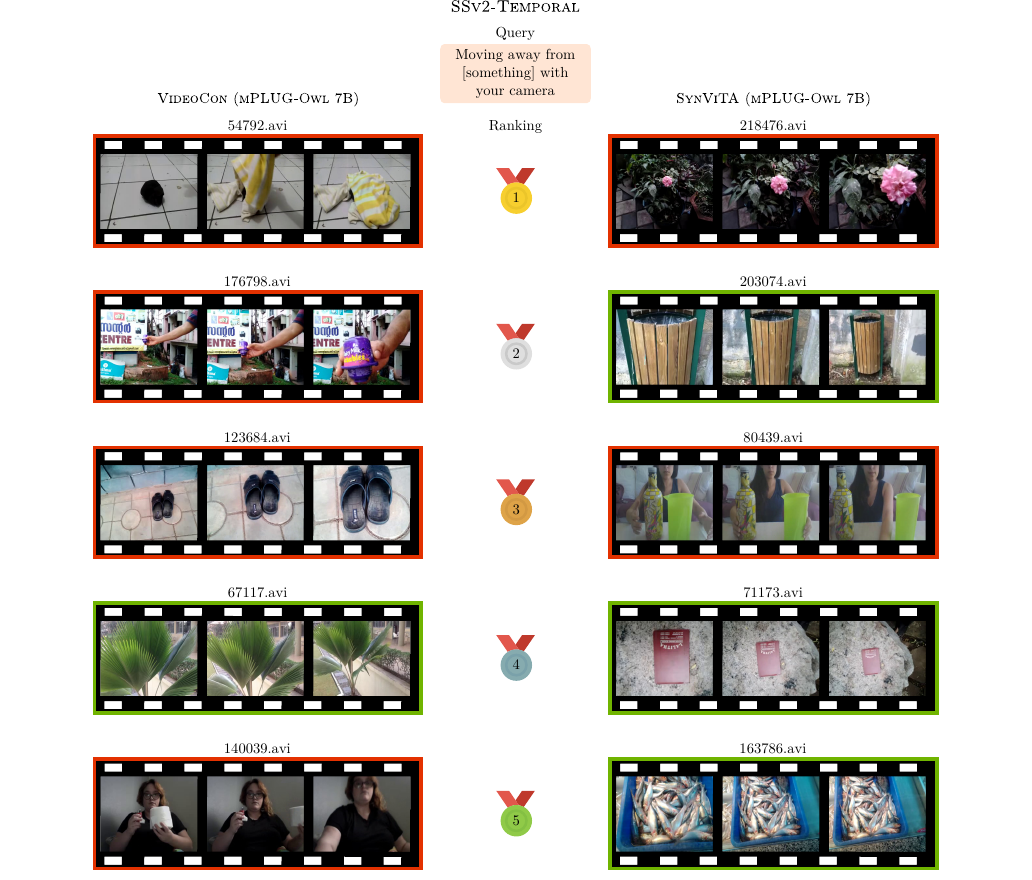}
    \caption{Comparison of rankings based on video-language alignment scores for the text-to-video retrieval task on SSv2-Temporal, using \methodshort (\mplugowl) against the baseline \videocon (\mplugowl) trained without synthetic videos. Videos marked with {\color{forestgreen(web)}green} borders correctly match the input text query, while those marked with {\color{red}red} borders do not.}
    \label{fig:ssv2_temporal_mplugowl}
\end{figure*}

\begin{figure*}[t!]
    \centering
    \includegraphics[width=\linewidth]{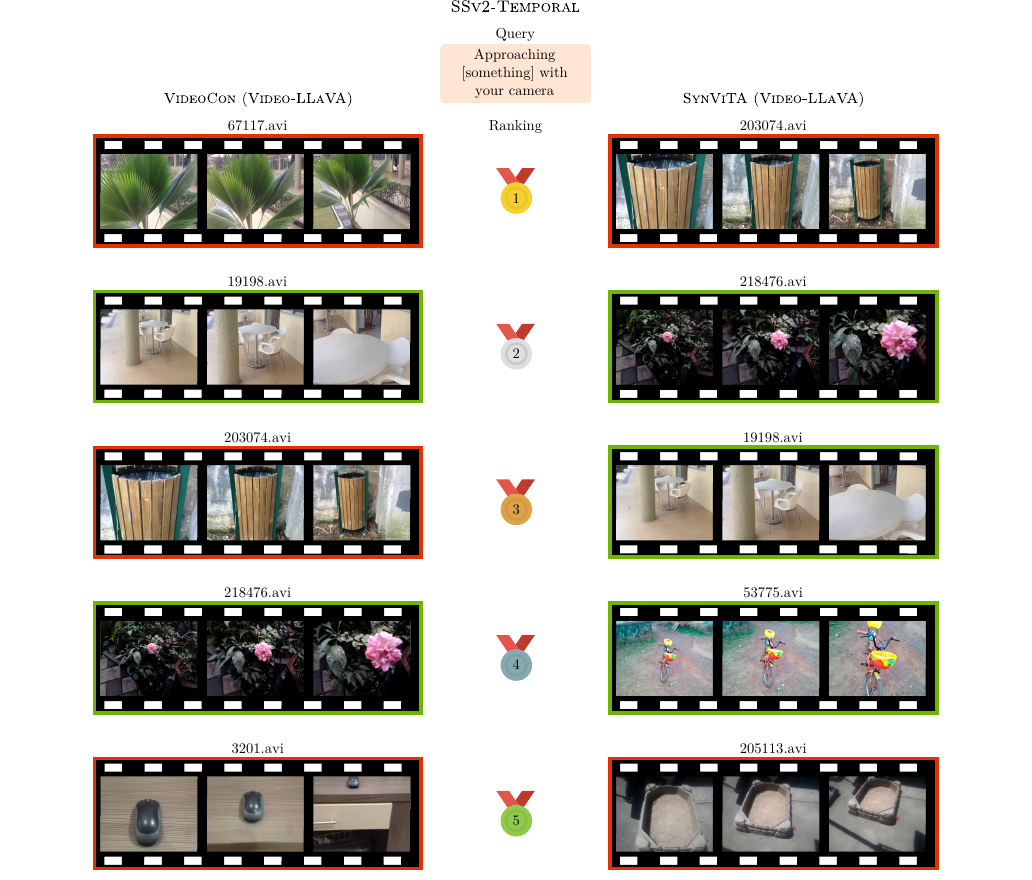}
    \caption{Comparison of rankings based on video-language alignment scores for the text-to-video retrieval task on SSv2-Temporal, using \methodshort (\videollava) against the baseline \videocon (\videollava) trained without synthetic videos. Videos marked with {\color{forestgreen(web)}green} borders correctly match the input text query, while those marked with {\color{red}red} borders do not.}
    \label{fig:ssv2_temporal_videollava}
\end{figure*}

\begin{figure*}[t!]
    \centering
    \includegraphics[width=\linewidth]{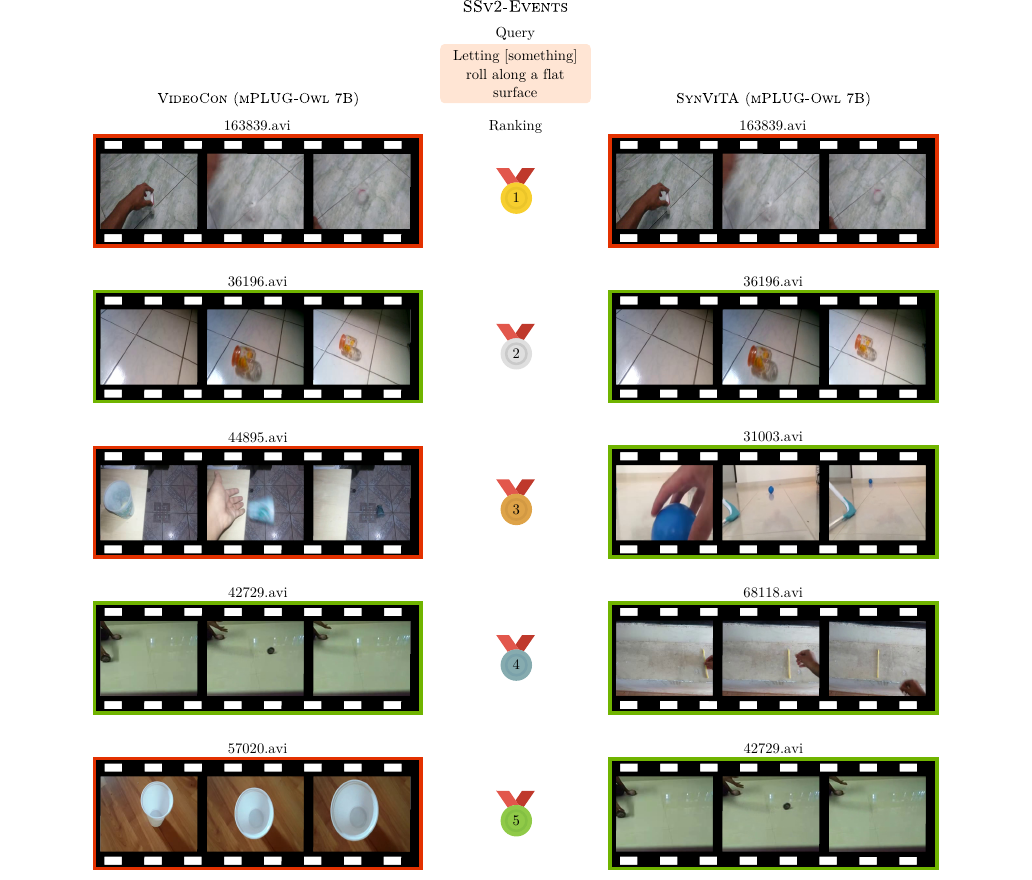}
    \caption{Comparison of rankings based on video-language alignment scores for the text-to-video retrieval task on SSv2-Events, using \methodshort (\mplugowl) against the baseline \videocon (\mplugowl) trained without synthetic videos. Videos marked with {\color{forestgreen(web)}green} borders correctly match the input text query, while those marked with {\color{red}red} borders do not.}
    \label{fig:ssv2_events_mplugowl}
\end{figure*}

\begin{figure*}[t!]
    \centering
    \includegraphics[width=\linewidth]{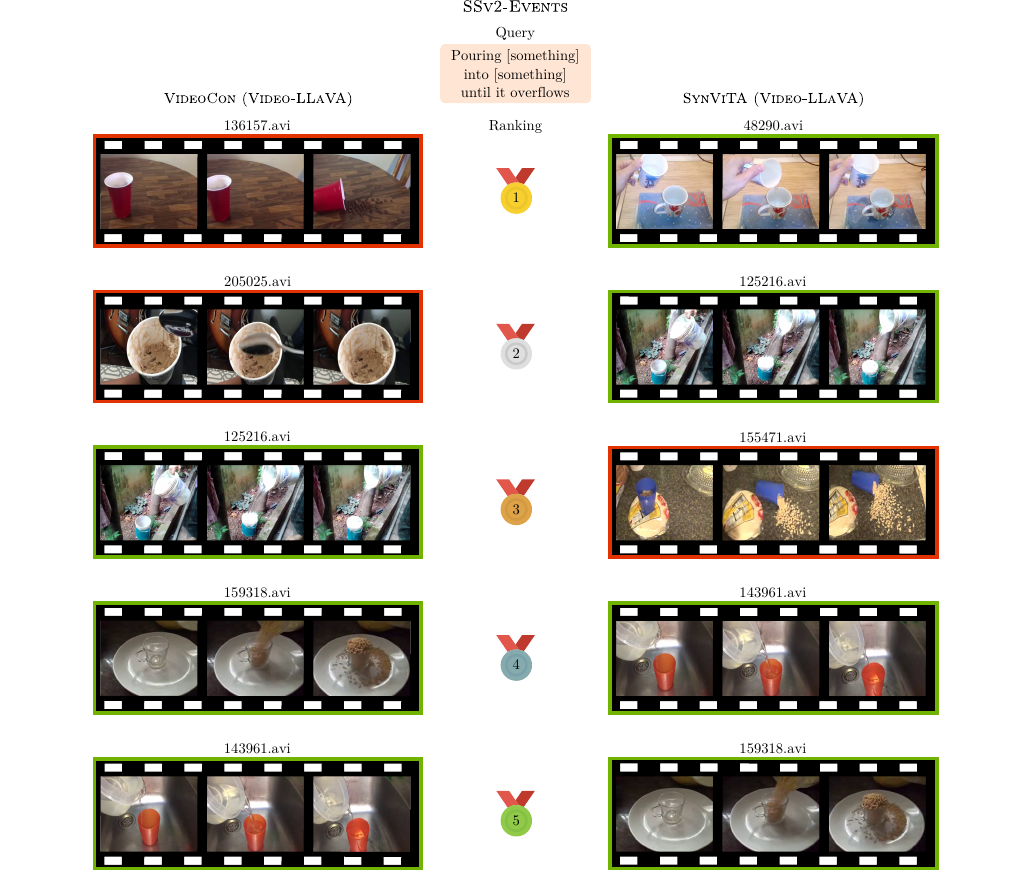}
    \caption{Comparison of rankings based on video-language alignment scores for the text-to-video retrieval task on SSv2-Events, using \methodshort (\videollava) against the baseline \videocon (\videollava) trained without synthetic videos. Videos marked with {\color{forestgreen(web)}green} borders correctly match the input text query, while those marked with {\color{red}red} borders do not.}
    \label{fig:ssv2_events_videollava}
\end{figure*}

\cref{fig:videocon_llm,fig:videocon_human} show examples of video-language alignment scores assigned by \methodshort (\mplugowl) and \methodshort (\videollava), compared to baselines trained without synthetic videos, for the video-language entailment task on \videocon LLM and \videocon Human and Human Hard, respectively. Similarly, \cref{fig:atphard} presents alignment scores for the video question answering task on ATP-Hard. Finally, \cref{fig:ssv2_temporal_mplugowl,fig:ssv2_temporal_videollava,fig:ssv2_events_mplugowl,fig:ssv2_events_videollava} show rankings based on video-language alignment scores for the text-to-video retrieval task on SSv2-Temporal and SSv2-Events, using \methodshort (\mplugowl) and \methodshort (\videollava) against the same baselines. %
\cref{fig:videocon_llm,fig:videocon_human} show some success cases on the video-language entailment task, where \methodshort assigns higher alignment scores to captions matching the videos compared to their negative counterparts (\eg, it better distinguishes the action of standing from sitting on both videos from \videocon Human). \cref{fig:videocon_llm} (rows 2 and 4) also shows two failure cases where \methodshort incorrectly associates videos with negative captions, unlike the baseline, which makes the correct associations. In the first case, \methodshort misjudges the number of children because the second child appears briefly at the end, looks similar to the first, and is never seen together. In the second case, \methodshort fails to detect that oil, not water, is added to the pan, likely because the oil is already inside the pan when the video starts. For the video question answering task (\cref{fig:atphard}), it better associates the scenario of a child sitting on its father's stomach versus its shoulders. Finally, on the text-to-video retrieval task, it better recognizes certain actions such as moving relative to the camera (\cref{fig:ssv2_temporal_mplugowl,fig:ssv2_temporal_videollava}), rolling something (\cref{fig:ssv2_events_mplugowl}), and pouring something (\cref{fig:ssv2_events_videollava}).

\clearpage

\clearpage

\clearpage

\clearpage

\end{document}